\documentclass[11pt]{article}

\usepackage[final]{acl}

\usepackage{times}
\usepackage{latexsym}

\usepackage[T1]{fontenc}

\usepackage[utf8]{inputenc}

\usepackage{microtype}

\usepackage{inconsolata}
\usepackage{enumitem}
\usepackage{caption}

\usepackage{graphicx}

\usepackage{algorithm}
\usepackage{algorithmic}
\usepackage{booktabs}
\usepackage{subcaption}
\usepackage{amsmath,amsfonts,bm}
\usepackage{amsthm}
\usepackage{bbm}

\newcommand{\formattable}[2]{\begin{tabular}{@{}c@{}}#1\\[-2pt]{\color{gray}\tiny  $\pm$  #2}\end{tabular}}

\title{Demystifying Data Organization for Enhanced LLM Training}

\author{
\textnormal{Yalun Dai$^{1,2}$,
Yangyu Huang$^{2}$$^\dagger$$^{\diamond}$,
Tongshen Yang$^{1}$,
Yonghan Wang$^{1}$,
Xin Zhang$^{2}$,}
\\
Wenshan Wu$^{2}$,
Qihao Zhao$^{1}$,
Hao Li$^{1}$,
Yuanyuan Gao$^{3}$,
Kim-Hui Yap$^1$$^\dagger$,
Scarlett Li$^2$
\vspace{3mm}
\\
$^1$Nanyang Technological University \quad
$^2$Microsoft Research  \\
$^3$The Hong Kong University of Science and Technology \quad
\vspace{-3mm}
}

\begin{document}
\maketitle
\begin{abstract}
Large Language Models (LLMs) have revolutionized various fields, yet their training efficiency is heavily reliant on effective data curation.
While data selection has been widely studied, the strategic data organization for enhanced training remains an underexplored area, particularly since current LLMs are often trained for only one or a few epochs.
This paper systematically explores the influence of data organization on LLM training by reusing pre-computed sample-level scores originally generated for data efficiency, thereby incurring minimal additional computational overhead.
We identify and formalize four key guidances for optimizing data organization: Boundary Sharpening, Cyclic Scheduling, Curriculum Continuity, and Local Diversity.
Guided by them, we introduce two novel data ordering methods termed STR and SAW.
Extensive experiments across different model scales and data sizes, encompassing both pre-training and SFT stages, validate the effectiveness of our summarized guidances.
They also demonstrate the robustness of our proposed data ordering methods in enhancing the stability and performance of LLM training.
Github Link: \url{https://github.com/microsoft/data-efficacy/}
\end{abstract}

{
  \renewcommand{\thefootnote}%
    {}
  \footnotetext[2]{
   Work done during the internships of Yalun Dai at Microsoft Research.
  \noindent$^{\diamond}$Project leader.
  \noindent$^\dagger$Corresponding Authors.
  }
}

\vspace{-0.5cm}
\section{Introduction}
Large Language Models (LLMs) have revolutionized various fields by providing unprecedented capabilities in areas such as coding \citep{luo2023wizardcoder, yu2024wavecoder}, science \citep{van2023ai}, nature \citep{huang2025peace}, and healthcare \citep{vaananen2021ai}.
While data curation strategies, including acquisition \citep{trafilatura2021acquisition,bevendorff2023acq,chang2024redstone}, mixing \citep{albalak2023efficientmix,ge2025bimixbivariatedatamixing,ye2024datamix}, synthesis \citep{chen2025advancingsyn,zhou2024jiuzhang3syn}, deduplication \citep{abbas2023semdedup,tirumala2023d4}, filtering \citep{li2024superfiltering,goyal2024filter,thrush2024filter}, and selection \citep{albalak2024dssurvey,xie2023data,dai2025training,gu2025data}, have been extensively studied and significantly advanced LLM development. 

However, a critical but underexplored aspect of LLM training efficacy lies \textit{not merely in what data is used, but how it is presented in a given dataset.}

\begin{figure}[!tb]
  \centering
   \includegraphics[width=1.0\columnwidth]{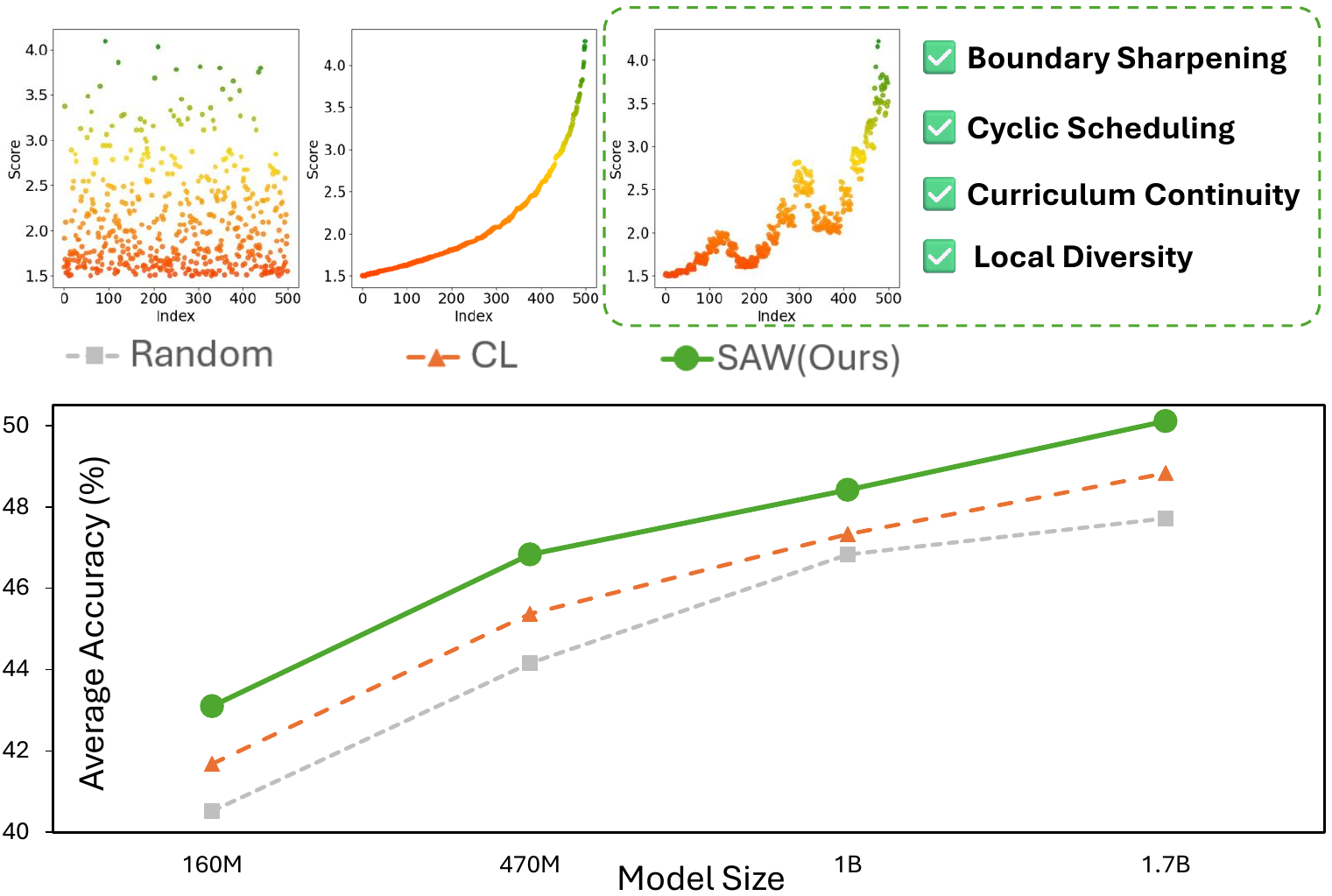}
    \caption{Performance comparison of different data organization strategies.
Bottom: Average accuracy (\%) under increasing model sizes (160M–1.7B).
Our proposed SAW consistently outperforms Random and Curriculum Learning (CL) across all scales, with larger gains at higher capacities.
Top: Score–index distributions of training data organized by different strategies.
Compared to Random and CL, SAW induces more structured and progressive score patterns, revealing increasingly organized curricula.}
  \label{fig:cover}
\vspace{-0.3cm}
\end{figure}

Despite the existence of well-curated datasets, the strategic organization of these training samples remains a significant challenge. 
Current LLMs are often trained for only one or a few epochs over massive corpora \citep{openAI2024,llama32024llama,gemini2023gemini,qwen2025qwen3}. Under such paradigms, the temporal order in which training samples are presented becomes a first-order factor shaping the learning trajectory.
Existing approaches like Curriculum Learning \citep{bengio2009cl}, which arranges data samples by difficulty, and DELT \citep{dai2025dataefficacy}, which employs a folding learning strategy to look back at the similar scoring distribution of training data and offer valuable insights into data organization. 
Yet, none of these works provide a systematic exploration of comprehensive guidance specifically tailored for data ordering in LLM training.

To bridge this critical gap, this paper presents the first systematic exploration into the influence of data organization on LLM training.
A key innovation of our approach is the reuse of pre-computed, sample-level scores, originally generated for data efficiency \citep{lozhkov2024fineweb-edu,wettig2024qurating}, which incurs almost zero additional computational overhead.
Through extensive analysis, we identify and formalize four novel and critical guidances for optimizing data organization:

\begin{itemize}[left=0cm]
\item \textbf{Boundary Sharpening.}
Start training with low-score data to stabilize early learning, and finish with high-score data to enhance the model's performance. This approach helps the model handle complex reasoning tasks effectively.
%
\item \textbf{Cyclic Scheduling.}
Instead of a strict low-to-high score curriculum that can cause forgetting, cyclic scheduling periodically reintroduces a mix of data distribution. This helps the model retain fundamental knowledge and prevents forgetting.
%
\item \textbf{Curriculum Continuity.}
Smooth transitions between different score distributions of data prevent disruptions in training, maintaining stability and ensuring the model adapts progressively without losing momentum.
%
\item \textbf{Local Diversity.} 
Mixing diverse samples within mini-batches prevents overfitting and promotes learning of generalizable features, improving the model's ability to generalize from the data.
\end{itemize}

To demonstrate how these guidances can be employed in practice, we instantiate them into two novel ordering strategies termed STR and SAW, which systematically integrate these guidances to create an optimized training sequence.
The score-index distributions of these organized sequences are presented in the bottom part of Figure~\ref{fig:cover}.
Extensive experiments across diverse model scales and data sizes, encompassing both pre-training and supervised fine-tuning (SFT) stages, confirm the effectiveness of our formalized guidances.
The top part of Figure~\ref{fig:cover} illustrates the efficacy of the proposed STR and SAW methods in improving LLM training stability and overall performance.

\section{Problem Formulation}

Training large language models (LLMs) with vast datasets from the web is costly.
To improve data efficiency, previous methods spend a lot of resources calculating scores for each data sample (e.g., quality, difficulty, learnability, educational value), but use them just once for selecting data.
In contrast, this work aims to reuse these scores to organize training data, improving model performance.
We formalize this process in three stages:

\textbf{Data Scoring.}
This stage evaluates the raw dataset $\mathcal{D}=\{x_1, \dots, x_{|\mathcal{D}|}\}$ to derive fine-grained metrics. Let $g$ denote the scoring function that maps $\mathcal{D}$ to a score vector $\bm{\gamma} \in \mathbb{R}^{|\mathcal{D}|}$, representing specific criteria such as quality or difficulty:
\begin{equation}
\small
    \bm{\gamma} = g(\mathcal{D}) = [\gamma_1, \gamma_2, \dots, \gamma_{|\mathcal{D}|}]^\top.
    \label{eq:gamma}
\end{equation}

\textbf{Data Selection.}
This process identifies an optimal subset to reduce training costs. Let $f_s$ be the selection function with a selection ratio $R$. 
It filters $\mathcal{D}$ based on $\bm{\gamma}$, retaining the top-$K$ samples where $K = \lfloor R \cdot |\mathcal{D}| \rfloor$. The resulting subset $\mathcal{D}_{\text{sub}}$ changes the dataset scale but does not inherently determine the training order:
\begin{equation}
\small
    \mathcal{D}_{\text{sub}} = f_s(\mathcal{D}; \bm{\gamma}, K) = \left\{ x_i \in \mathcal{D} \mid r(\gamma_i) \le K \right\},
    \label{eq:selection}
\end{equation}
where $r(\cdot)$ indexes elements in descending order of their scores $\bm{\gamma}$.

\textbf{Data Organization.}
This is the core part of our work, focusing on the strategic arrangement of data.
Unlike selection, Data Organization $f_o$ does not alter the size but reorganizes the sequence of the dataset based on a permutation $\pi$ derived from the scores $\bm{\gamma}$.
As a result, the initially disordered data is transformed into ordered data:
\begin{equation}
\small
    \mathcal{D}_{\text{ord}} = f_o(\mathcal{D}; \bm{\gamma}) = [x_{\pi(1)}, x_{\pi(2)}, \dots, x_{\pi(K)}].
    \label{eq:organization}
\end{equation}
where the permutation $\pi$ effectively translates the score into an ordering index for each data point. 
Finally, the overall pipeline is formalized as $\mathcal{D}_{\text{train}} = f_o(f_s(\mathcal{D}; \bm{\gamma}, K); \bm{\gamma})$, $f_o$ operates on either the full dataset or a selected subset.
To maximize the utility of the pre-computed scores, our pipeline employs the \textit{identical} score vector $\bm{\gamma}$ to guide both data selection and organization, ensuring the computational effort in scoring is used without additional overhead.
As a special case of data organization, Curriculum Learning (CL) corresponds to an ordering function $f_o$ that sorts samples in ascending order of scores $\bm{\gamma}$, yielding $\mathcal{D}_{\text{sort}}$.

\section{Guidances for Data Organization}
\label{sec:method_guidance}
\subsection{G1: Boundary Sharpening.} \label{sec:method_guidance1}
\textbf{Motivation.}
The training trajectory of LLMs is highly sensitive to data distribution at the optimization boundaries, particularly at the start and end of training.
Recent findings indicate that improper data characteristics at initialization can destabilize early optimization and hinder convergence \citep{2023boundary,kalra2024warmupboundary,paul2022lotteryboundary,zhang2025beyond}.
Likewise, the data distribution at the end of training greatly affects the model's attainable performance.
Insufficient quality or complexity restricts performance on downstream tasks \citep{hu2024minicpmboundary,llama32024llama}.
Thus, controlling data scoring distribution at these phases is crucial for optimal training dynamics.

\noindent
\textbf{Guidance.}
Boundary Sharpening advocates explicitly controlling the data characteristics at the start and end of training.

\vspace{-0.2cm}
\begin{algorithm}[htp]
    \captionsetup{font=small}
    \caption{Segment Ordering (SEG)}
    \label{alg:seg}
    \small
    \begin{algorithmic}[1]
    \item[\textbf{Input:}] Sorted dataset $\mathcal{D}_\text{sort}$, intervals $\{I_l\}_{l=0}^{L-1}$
    
    \item[\textbf{Output:}] Organized dataset $\mathcal{D}_{\text{ord}}$
    
    \item[\textbf{Desc.:}] Percentile-based partitioning \& shuffling
    \STATE $\{\mathcal{D}_l\}_{l=0}^{L-1} \gets \emptyset$
    \FORALL{$x$ in $\mathcal{D}_{\text{sort}}$}
        \STATE $k \sim \text{Uniform}(\{l \mid \text{rank}(x) \in I_l\})$
        \STATE $\mathcal{D}_k \leftarrow \mathcal{D}_k \cup \{x\}$
    \ENDFOR
    \STATE $\{\tilde{\mathcal{D}_l}\}_{l=0}^{L-1} \leftarrow \text{Shuffle} \{\mathcal{D}_l\}_{l=0}^{L-1}$
    
    \STATE $\mathcal{D}_{\text{ord}} \leftarrow \text{Concatenate}(\tilde{\mathcal{D}}_0,\tilde{\mathcal{D}}_1, \dots, \tilde{\mathcal{D}}_{L-1})$

    \end{algorithmic}
\end{algorithm}
\vspace{-0.4cm}

\noindent
\textbf{Implementation (Alg. \ref{alg:seg}).}
To instantiate Boundary Sharpening in practice, we propose a \textbf{Segment Ordering (SEG)} strategy that discretizes the training sequence into $L$ distinct segments, $\mathcal{D}_0, \dots, \mathcal{D}_{L-1}$, derived from the sorted data $\mathcal{D}_\text{sort}$.
Each segment $\mathcal{D}_l$ is defined by a target percentile interval $[p_{\text{start}}^{(l)}, p_{\text{end}}^{(l)}]$ of $\mathcal{D}_\text{sort}$.
To construct the curriculum, we map samples to segments based on their score rankings.
To prevent duplication in overlapping regions, shared samples are evenly distributed among segments.
Each segment $\mathcal{D}_l$ is then randomly shuffled to create $\tilde{\mathcal{D}}_l$.
The final organized dataset is constructed by concatenating these processed segments:
$\mathcal{D}_{\text{ord}} = [\tilde{\mathcal{D}}_0,\tilde{\mathcal{D}}_1, \dots, \tilde{\mathcal{D}}_{L-1}]$.

\subsection{G2: Cyclic Scheduling.}
\label{sec:method_guidance2}

\textbf{Motivation.} While a strict scoring-based sorting is intuitive, it poses risks in the one-pass training regime typical of LLMs \citep{tay2019simpleonepass,tudor2016hardonepass}.
As the model transitions exclusively to complex samples in later stages, it tends to forget fundamental knowledge acquired from early simple data \citep{wang2021clsurvey,dai2025dataefficacy}.
Standard solutions like Baby Step methods \citep{spitkovsky2010babystep,cirik2016visualizingbabystep}, which mix previous simple data with new hard data, are often impractical in LLM training as they require explicit data replay, inflating the training costs.

\noindent
\textbf{Guidance.}
Cyclic Scheduling advocates periodically revisiting training samples across the full score spectrum during one-pass training.

\vspace{-0.2cm}
\begin{algorithm}[htp]
    \small
    \captionsetup{font=small}
    \caption{Folding Ordering (FO)}
    \label{alg:folding}
    \begin{algorithmic}[1]
        \item[\textbf{Input:}] Sorted dataset $\mathcal{D}_\text{sort}$, folding layers $L$
        \item[\textbf{Output:}]  Organized dataset $\mathcal{D}_\text{ord}$
        \item[\textbf{Desc.:}]  Strided partitioning of sorted data
        
            \FOR{$l = 0$ \TO $L-1$}
            \STATE $\mathcal{D}_l \leftarrow \{ \mathcal{D}_{\text{sort}}[i] \mid 0 \le i \le |\mathcal{D}_{\text{sort}}|-1, \ i \equiv l \pmod L \}$
        \ENDFOR
        
        \STATE $\mathcal{D}_\text{ord} \leftarrow \text{Concatenate}(\mathcal{D}_0, \mathcal{D}_1, \dots, \mathcal{D}_{L-1})$
    \end{algorithmic}
\end{algorithm}
\vspace{-0.4cm}


\noindent
\textbf{Implementation (Alg. \ref{alg:folding}).} 
To practically instantiate Cyclic Scheduling, we introduce
a \textbf{Folding Ordering (FO)} strategy based on pre-computed scores $\bm{\gamma}$.
Let $\pi_{\text{sort}}$ be the permutation indices of sorted dataset $\mathcal{D}_\text{sort}$, such that $\gamma_{\pi_{\text{sort}}(0)} \le \dots \le \gamma_{\pi_{\text{sort}}(|D|-1)}$.
We divide the training dataset into $L$ folding layers. 
To ensure each folding cycle covers the entire score spectrum, we distribute the sorted samples in a strided partitioning way. 
Specifically, the dataset is reorganized into a series of cycles $\mathcal{D}_0, \dots, \mathcal{D}_{L-1}$.
Each cycle corresponds to a folding layer in the training sequence, and the $l$-th folding layer $\mathcal{D}_l$ ($0 \le l < L$) collects samples with sorted ranks congruent to $l$ modulo $L$:
\begin{equation}
\scriptsize
    \mathcal{D}_l = \left[ x_{\pi_{\text{sort}}(i)} \mid i \equiv l \pmod L, \ 0 \le i \le |\mathcal{D}|-1 \right].
    \label{eq:cyclic_buckets}
\end{equation}
The final ordered dataset $\mathcal{D}_\text{ord}$ is created by concatenating these cycles: $\mathcal{D}_\text{ord} = [\mathcal{D}_0, \mathcal{D}_1, \dots, \mathcal{D}_{L-1}]$.
Within each cycle $\mathcal{D}_l$, samples keep their relative low-to-high score order.
This construction effectively transforms a globally increasing trajectory into a periodic pattern, allowing the model to revisit early basic concepts regularly while progressively tackling samples with higher scores.
%

\subsection{G3: Curriculum Continuity.}
\label{sec:method_guidance3}

\textbf{Motivation.} The dynamics of the optimization process are highly sensitive to the sequence of data presentation.
Abrupt fluctuations in data attributes may induce shocks to the optimizer, manifesting as rapid switching between relatively low and high-scoring samples \citep{kong2021adaptivecontinuity,weinshall2018curriculumcontinuity}.
%
Such shifts often lead to high variance in gradient estimates, disrupting the optimization stability and potentially causing loss divergence.

\noindent
\textbf{Guidance.}
Curriculum Continuity advocates maintaining smooth transitions in data attributes throughout the training sequence.

\vspace{-0.2cm}
\begin{algorithm}[htp]
    \captionsetup{font=small}
    \caption{Zig-zag Ordering (ZIG)}
    \label{alg:zigzag}
    \small
    \begin{algorithmic}[1]
        \item[\textbf{Input:}] Sorted dataset $\mathcal{D}_\text{sort}$, folding layers $L$
        \item[\textbf{Output:}] Organized dataset $\mathcal{D}_\text{ord}$
        \item[\textbf{Desc.:}] Enhancing continuity across partitions
        
        \STATE $\{\mathcal{D}_l\}_{l=0}^{L-1} \leftarrow \text{FO}(\mathcal{D}_\text{sort}, L)$ \COMMENT{See Alg. \ref{alg:folding}}
        \FOR{$l = 0$ \TO $L-1$}
            \IF{$l$ is odd}
                \STATE $\mathcal{D}_l \leftarrow \text{Reverse}(\mathcal{D}_l)$
            \ENDIF
        \ENDFOR
        
        \STATE $\mathcal{D}_\text{ord} \leftarrow \text{Concatenate}(\mathcal{D}_0, \mathcal{D}_1, \dots, \mathcal{D}_{L-1})$
    \end{algorithmic}
\end{algorithm}
\vspace{-0.4cm}


\noindent
\textbf{Implementation. (Alg. \ref{alg:zigzag})} 
To instantiate Curriculum Continuity in practice, we propose the \textbf{Zig-zag Ordering (ZIG)} strategy. 
FO (Sec. \ref{sec:method_guidance2}) causes sharp attribute drops when switching between cycles, violating the continuity required for stable optimization. 
ZIG resolves this by simply reversing the sample order in every odd-indexed cycle (i.e., $\mathcal{D}_1, \mathcal{D}_3, \dots$) from Eq. \ref{eq:cyclic_buckets}.
The final dataset $\mathcal{D}_\text{ord}$ concatenates these partially reversed subsets to a continuous triangle-wave pattern, ensuring that adjacent cycles typically connect at points of similar score and yielding a smoother training trajectory.

\subsection{G4: Local Diversity.}
\label{sec:method_guidance4}
\textbf{Motivation.} 
While establishing a global curriculum (e.g., via cyclic scheduling or curriculum continuity) is beneficial, strictly sorting the dataset by scores $\bm{\gamma}$ negatively impacts \textit{Local Diversity}.
When training samples within a mini-batch possess nearly identical scores and attributes, the diversity of gradients diminishes \citep{jiang2014selfdiversity}.
This lack of intra-batch variance may bias the optimizer towards specific, repetitive patterns, leading to overfitting and reducing the model's ability to learn generalizable features \citep{yin2018gradientdiversity}.
%

\noindent
\textbf{Guidance.}
Local Diversity advocates preserving sufficient heterogeneity among training samples within local sections of the training sequence.

\vspace{-0.2cm}
\begin{algorithm}[htp]
    \captionsetup{font=small}
    \caption{Jittering Ordering (JIT)}
    \label{alg:jitter}
    \small
    \begin{algorithmic}[1]
        \item[\textbf{Input:}] Sorted dataset $\mathcal{D}_\text{sort}$, window size $w$
        \item[\textbf{Output:}] Organized dataset $\mathcal{D}_\text{ord}$
        \item[\textbf{Desc.:}] Injecting index noise for local diversity
        \STATE $L \leftarrow \lceil |\mathcal{D}_\text{sort}| / w \rceil$
    
        \FOR{$l = 0$ \TO $L-1$}
            \STATE $\mathcal{D}_l \leftarrow \text{Shuffle}(\mathcal{D}_{\text{sort}}[l \cdot w : (l+1) \cdot w])$
        \ENDFOR
        
        \STATE $\mathcal{D}_\text{ord} \leftarrow \text{Concatenate}(\mathcal{D}_0, \mathcal{D}_1, \dots, \mathcal{D}_{L-1})$
    \end{algorithmic}
\end{algorithm}
\vspace{-0.4cm}


\noindent
\textbf{Implementation. (Alg. \ref{alg:jitter})}
To instantiate Local Diversity in practice, we introduce \textbf{Jittering Ordering (JIT)} strategy by injecting tunable noise into the strictly sorted data $\mathcal{D}_\text{sort}$ (e.g., CL, FO, and ZIG).
%
Specifically, we partition the sorted dataset $\mathcal{D}_{\text{sort}}$ into contiguous buckets of size $w$ (diversity window).
Within each bucket, we apply random shuffling while preserving the relative order between buckets.
The final dataset $\mathcal{D}_\text{ord}$ concatenates these locally shuffled buckets, restoring essential gradient noise while maintaining the global trend.
%
%
%

\section{Methodology}
\label{sec:cross-guidance}
Besides discussing each guidance individually, we also propose several cross-guidance strategies to explore their interplay (Alg. \ref{alg:sec_saw}).

\textbf{Cross-guidance strategies.}
A cross-guidance strategy combines multiple guidances into one data ordering function by assigning different ordering mechanisms to different parts of the training sequence.
This approach enables controlled composition of global progression, periodic review, continuity, and local diversity within a unified framework, rather than relying on a single ordering heuristic.

\begin{figure}[hbt]
  \centering
  \includegraphics[width=1.0\columnwidth]{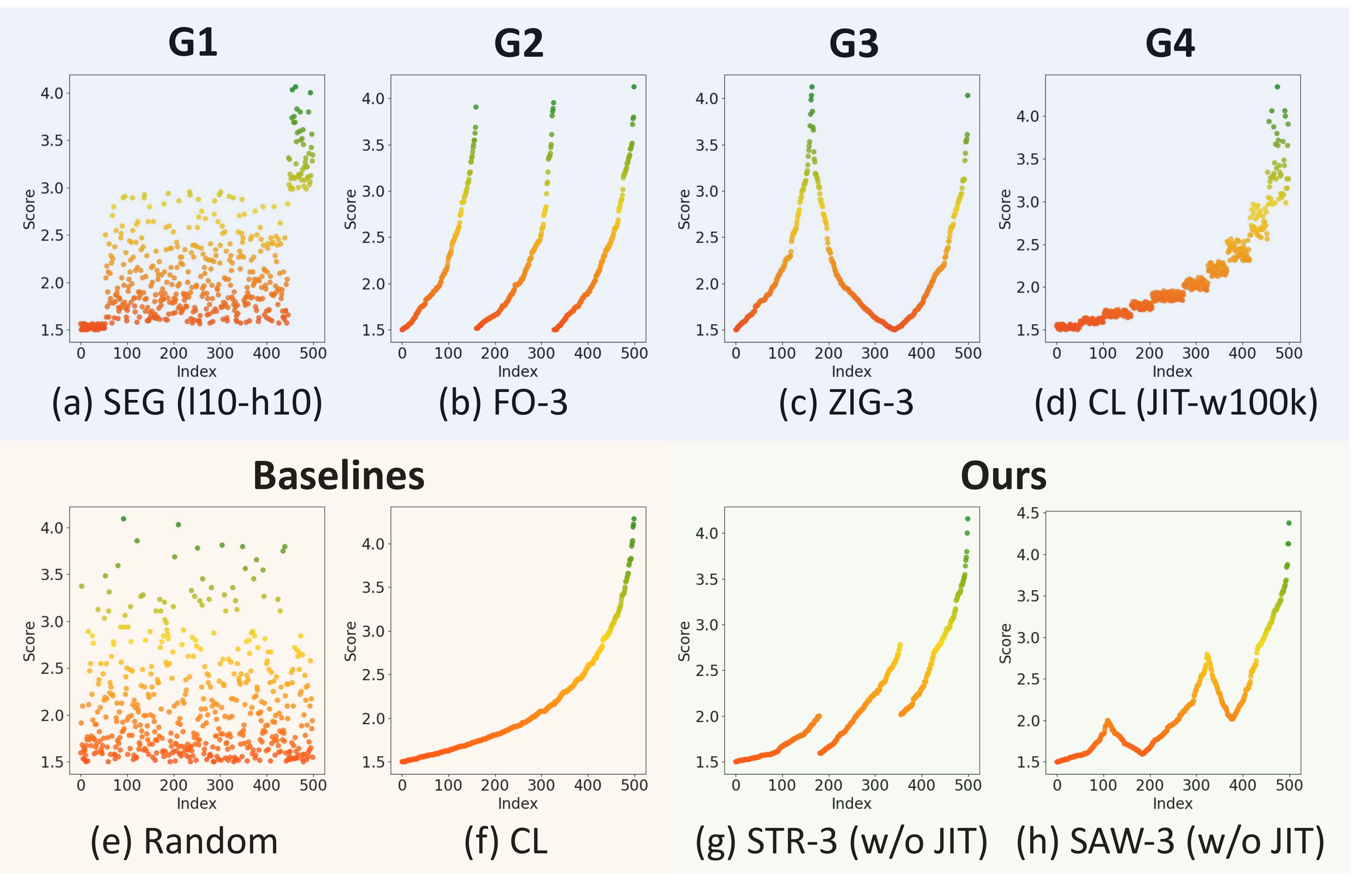}
  \caption{Visualization of score-index distribution under different data organization strategies. For clear comparison, $L=3$ is used for FO, ZIG, STR, and SAW.}
  \label{fig:vis_sort_distribution}
\vspace{-0.1cm}
\end{figure}

\vspace{-0.2cm}
\begin{algorithm}[htp]
    \captionsetup{font=small}
    \caption{STR \& SAW}
    \label{alg:sec_saw}
    \small
    \begin{algorithmic}[1]
        \item[\textbf{Input:}] Sorted dataset $\mathcal{D}_\text{sort}$, split points $\{p_l\}_{l=1}^{K-1}$, radius $\rho$, window size $w$, folding layers $L$, mode $T \in \{\text{STR}, \text{SAW}\}$
        \item[\textbf{Output:}] Organized dataset $\mathcal{D}_\text{ord}$
        
        \item[\textbf{Desc.:}] Cross-guidance strategy. 
        \STATE $\{\mathcal{D}^{t}_l\}_{l=1}^{K-1} \leftarrow \{ \mathcal{D}_{\text{sort}}[p_l - \rho : p_l + \rho] \}$
        \STATE $\{\mathcal{D}^{s}_l\}_{l=0}^{K-1} \leftarrow \{ \mathcal{D}_{\text{sort}}[p_l + \rho : p_{l+1} - \rho] \}$
        \item[] $\text{with } p_0 = -\rho, p_K = |\mathcal{D}_{\text{sort}}| + \rho$
        
        \FOR{each $\mathcal{D}^{t}_l \in \{\mathcal{D}^{t}_l\}$}
            \IF{$T = \text{STR}$}
                \STATE $\mathcal{D}^{t}_l \leftarrow \text{FO}(\mathcal{D}^{t}_l,L)$ \COMMENT{G2, see Alg. \ref{alg:folding}}
            \ELSIF{$T = \text{SAW}$}
                \STATE $\mathcal{D}^{t}_l \leftarrow \text{ZIG}(\mathcal{D}^{t}_l,L)$ \COMMENT{G3, see Alg. \ref{alg:zigzag}}
            \ENDIF
        \ENDFOR
        
        \STATE $\mathcal{D}_\text{ord} \leftarrow \text{Concatenate}(\mathcal{D}^{s}_0, \mathcal{D}^{t}_1, \mathcal{D}^{s}_1, ..., \mathcal{D}^{t}_{K-1}, \mathcal{D}^{s}_{K-1})$
        
        \IF{$w > 0$}
            \STATE $\mathcal{D}_\text{ord} \leftarrow \text{JIT}(\mathcal{D}_\text{ord}, w)$ \COMMENT{G4, see Alg. \ref{alg:jitter}}
        \ENDIF
    \end{algorithmic}
\end{algorithm}
\vspace{-0.2cm}

\textbf{Stair Ordering (STR).}
This strategy follows guidances of \textbf{G1, G2, and G4}. 
To explore the benefits of global progression (G1) and periodic review (G2), STR maintains a global trend while injecting controllable folding mechanisms at local transitions to facilitate early knowledge review.
Building upon this, we optionally apply the JIT mechanism to data samples to enhance local diversity (G4).

%
Specifically, we partition the $\mathcal{D}_{\text{sort}}$ into $K$ sections by identifying $K-1$ split points. 
Around each point, we define a transition region $\mathcal{D}^{t}$ with a radius $\rho$, while the remaining portions constitute stable regions $\mathcal{D}^{s}$. 
In the final sequence, stable regions retain their monotonic order, whereas transition regions apply the interleaving logic $f_{\text{FO}}$ (Eq. \ref{eq:cyclic_buckets}) to facilitate knowledge review.

\textbf{Saw Ordering (SAW).}
SAW generalizes STR by integrating \textbf{G1, G2, G3, and G4}, and explicitly enforces continuity constraints during region transitions.
While the STR ordering strategy effectively integrates G1\&2, the use of folding mechanisms to transition between regions introduces abrupt shifts in data attributes, posing a challenge to G3 described in Sec. \ref {sec:method_guidance3}. 

To address this and explore the benefits of attribute continuity, SAW refines the STR framework by replacing the folding $f_{\text{FO}}$ within transition regions with the Zig-zag mechanism $f_{\text{ZIG}}$, thereby ensuring a smoother transition between samples at region boundaries.
Similarly, we also optionally apply the JIT mechanism (G4).

%

\section{Experiment}
\subsection{Experimental Setup}
\textbf{Data.}
For general data, we report FineWeb-Edu \citep{lozhkov2024fineweb-edu} results in the main text and defer QuRatedPajama \citep{wettig2024qurating} to Appendix \ref{appendix:more_results}. 
For specific tasks, we utilize DeepMath-103K \citep{he2025deepmath} for math reasoning and OpenCodeInstruct \citep{ahmad2025opencodeinstruct} for code generation. See Appendix \ref{appendix:setup data} for dataset details.

\noindent\textbf{Models.} 
We apply the \textbf{Mistral} \citep{mistral} architecture for pre-training on general data, and the \textbf{Qwen3} \citep{qwen2025qwen3} for SFT in the math and code data using the official pre-trained weights.
See Appendix \ref{appendix:setup model} for model details.
%

\noindent\textbf{Baselines.}
We use random sampling (\textbf{Random}) and vanilla curriculum learning (\textbf{CL}) as baselines. 
Following Sec. \ref{sec:method_guidance}, we verify the effectiveness of each guidance through specific implementations: \textbf{SEG} (G1), \textbf{FO} (G2), \textbf{ZIG} (G3), and \textbf{JIT} (G4).
Besides, we investigate the cross-guidance strategies of the proposed \textbf{STR} and \textbf{SAW} in Sec. \ref{sec:cross-guidance}.
These strategies are visualized in Figure \ref{fig:vis_sort_distribution}.
See Appendix \ref{appendix:setup training} for training details.

For additional details of the experimental setup, such as training and evaluation, see Appendix \ref{appendix:setup_all}.

\vspace{-0.2cm}
\begin{table*}[htp]
    \centering
    \scriptsize
    \setlength{\tabcolsep}{3.2pt}
    \caption{Performance comparison of SEG variants and baselines to evaluate G1.}
    \label{tab:merged_results_g1}
\vspace{-0.3cm}
    
    \begin{tabular}{l | cccccccc |c | ccc |ccc}
        \toprule
        & \multicolumn{9}{c|}{\textbf{Pre-training} (FineWeb-Edu)} & \multicolumn{6}{c}{\textbf{SFT} (DeepMath-103K | OpenCodeInstruct)} \\
        \cmidrule(lr){2-10} \cmidrule(lr){11-16}
        & ARC-c & ARC-e & HS & LAMB & OBQA & PIQA & SciQ & Wino & Avg. & AIME24 & AIME25 &Avg. & HumanEval & MBPP &Avg.\\
        \midrule
        Random & \formattable{21.47}{0.18} & \formattable{37.50}{0.23} & \formattable{27.57}{0.15} & \formattable{15.97}{0.12} & \formattable{25.60}{0.20} & \formattable{57.80}{0.51} & \formattable{59.50}{0.48} & \formattable{51.13}{0.04} & \formattable{37.09}{0.08} & \formattable{1.29}{0.02} & \formattable{1.30}{0.02} & \formattable{1.30}{0.01} & \formattable{57.93}{0.60} & \formattable{52.80}{0.06} & \formattable{55.37}{0.30} \\
        \midrule
        SEG(h10) & 21.02 & 35.54 & 27.57 & \textbf{16.23} & 26.80 & \textbf{58.52} & 60.30 & 51.22 & 37.10 $\uparrow$ & 1.89 & \textbf{2.55} & 2.22 & \textbf{66.40} & 52.00 & 59.20 \\
        SEG(h90) & 21.27 & 35.75 & 28.01 & 15.40 & 25.80 & 58.09 & 58.80 & \textbf{51.30} & 36.93 $\downarrow$ & 2.10 & 1.82 & 1.96 & 59.10 & 53.20 & 56.15 \\
        SEG(l10) & 22.72 & \textbf{37.16} & 28.27 & 15.83 & \textbf{27.30} & 57.70 & 58.90 & 50.37 & 37.29 $\uparrow$ & 2.53 & 1.15 & 1.84 & 64.60 & \textbf{54.20} & \textbf{59.40} \\
        SEG(l90) & \textbf{23.38} & 36.53 & \textbf{28.58} & 15.58 & 26.80 & 58.00 & \textbf{61.20} & 50.20 & \textbf{37.53} $\uparrow$ & \textbf{2.64} & 2.29 & \textbf{2.47} & 59.10 & 53.00 & 56.05 \\
        \midrule
        SEG(h10-l10) & 21.63 & 35.57 & 27.69 & 16.21 & 27.10 & 58.69 & 60.20 & 50.43 & 37.12 $\uparrow$ & 2.32 & 2.03 & 2.18 & 60.30 & 53.40 & 56.85 \\
        SEG(l10-l10) & 22.11 & 36.03 & 28.10 & 14.54 & 27.00 & 58.55 & 58.10 & 51.94 & 36.92 $\downarrow$ & 2.05 & 1.67 & 1.86 & 63.40 & 52.60 & 58.00 \\
        SEG(l10-h10) & 22.61 & 37.75 & \textbf{28.78} & 15.08 & 28.40 & \textbf{58.76} & \textbf{62.20} & \textbf{52.64} & 38.28 $\uparrow$ & 1.89 & 1.61 & 1.75 & 59.10 & 53.40 & 56.25 \\
        SEG(h10-h10) & \textbf{22.95} & \textbf{38.59} & 28.42 & \textbf{16.61} & \textbf{29.00} & 58.54 & 61.90 & 50.59 & \textbf{38.33} $\uparrow$ & \textbf{2.37} & \textbf{2.14} & \textbf{2.26} & \textbf{66.40} & \textbf{54.60} & \textbf{60.50} \\

        \bottomrule
    \end{tabular}
\vspace{-0.2cm}
\end{table*}

\begin{table*}[h]
    \centering
    \scriptsize
    \setlength{\tabcolsep}{3.5pt}
    \caption{Performance comparison of FO variants and baselines to evaluate G2.}
    \label{tab:merged_results_guidance2}
\vspace{-0.3cm}
    
    \begin{tabular}{l | cccccccc|c | ccc | ccc}
        \toprule
        & \multicolumn{9}{c|}{\textbf{Pre-training} (FineWeb-Edu)} & \multicolumn{6}{c}{\textbf{SFT} (DeepMath-103K | OpenCodeInstruct)} \\
        \cmidrule(lr){2-10} \cmidrule(lr){11-16}
        & ARC-c & ARC-e & HS & LAMB & OBQA & PIQA & SciQ & Wino & Avg. & AIME24 & AIME25 & Avg. & HumanEval & MBPP & Avg.\\
        \midrule
        Random & \formattable{21.47}{0.18} & \formattable{37.50}{0.23} & \formattable{27.57}{0.15} & \formattable{15.97}{0.12} & \formattable{25.60}{0.20} & \formattable{57.80}{0.51} & \formattable{59.50}{0.48} & \formattable{51.13}{0.04} & \formattable{37.09}{0.08} & \formattable{1.29}{0.02} & \formattable{1.30}{0.02} & \formattable{1.30}{0.01} & \formattable{57.93}{0.60} & \formattable{52.80}{0.06} & \formattable{55.37}{0.30} \\
        CL     & 23.11 & 38.63 & 28.10 & 13.85 & \textbf{28.40} & 57.42 & 60.10 & 51.50 & 37.61 & 1.94 & 1.61 & 1.78 & 60.90 & 51.40 & 56.15 \\
        \midrule
        FO-2   & \textbf{24.23} & \textbf{38.97} & 28.43 & 14.65 & 28.20 & 58.16 & 62.30 & 50.59 & \textbf{38.19} $\uparrow$ & 1.89 & 1.56 & 1.73 & 60.90 & 53.60 & 57.25 \\
        FO-3   & 23.04 & 37.99 & 27.99 & 16.40 & 27.00 & \textbf{60.07} & 60.50 & 50.49 & 38.15 $\uparrow$ & \textbf{2.96} & \textbf{1.88} & \textbf{2.42} & 65.80 & 53.40 & 59.60 \\
        FO-4   & 23.22 & 37.31 & \textbf{28.77} & 16.15 & 27.20 & 58.37 & 59.00 & \textbf{51.80} & 37.71 $\uparrow$ & 2.16 & 1.72 & 1.94 & 62.10 & \textbf{55.60} & 58.85 \\
        FO-5   & 24.02 & 38.68 & 28.12 & 14.81 & 27.20 & 57.96 & \textbf{63.20} & 51.03 & 38.12 $\uparrow$ & 2.80 & 1.20 & 2.00 & \textbf{66.80} & 54.80 & \textbf{60.80} \\
        FO-20  & 21.60 & 37.11 & 27.55 & \textbf{16.83} & 26.80 & 57.34 & 59.70 & 50.03 & 37.06 $\downarrow$ & 2.42 & 1.61 & 2.02 & 60.90 & 51.20 & 56.05 \\
        FO-100 & 21.81 & 37.72 & 27.37 & 14.91 & 24.80 & 58.25 & 62.00 & 49.12 & 36.97 $\downarrow$ & 2.48 & 1.51 & 2.00 & 66.40 & 51.80 & 59.10 \\
        \bottomrule
    \end{tabular}
\vspace{-0.2cm}
\end{table*}

\begin{table*}[!h]
    \centering
    \scriptsize
    \setlength{\tabcolsep}{3.5pt}
    \caption{Performance comparison of ZIG variants and baselines to evaluate G3.}
    \label{tab:merged_results_guidance3}
\vspace{-0.3cm}
    
    \begin{tabular}{l | cccccccc | c | ccc | ccc}
        \toprule
        & \multicolumn{9}{c|}{\textbf{Pre-training} (FineWeb-Edu)} & \multicolumn{6}{c}{\textbf{SFT} (DeepMath-103K | OpenCodeInstruct)} \\
        \cmidrule(lr){2-10} \cmidrule(lr){11-16}
        & ARC-c & ARC-e & HS & LAMB & OBQA & PIQA & SciQ & Wino & Avg. & AIME24 & AIME25 & Avg. & HumanEval & MBPP & Avg.  \\
        \midrule
        Random & \formattable{21.47}{0.18} & \formattable{37.50}{0.23} & \formattable{27.57}{0.15} & \formattable{15.97}{0.12} & \formattable{25.60}{0.20} & \formattable{57.80}{0.51} & \formattable{59.50}{0.48} & \formattable{51.13}{0.04} & \formattable{37.09}{0.08} & \formattable{1.29}{0.02} & \formattable{1.30}{0.02} & \formattable{1.30}{0.01} & \formattable{57.93}{0.60} & \formattable{52.80}{0.06} & \formattable{55.37}{0.30} \\
        \midrule
        FO-2(or 5)  & \textbf{24.23} & \textbf{38.97} & 28.43 & 14.65 & \textbf{28.20} & 58.16 & \textbf{62.30} & \textbf{50.59} & 38.19 & 2.80 & 1.72 & 2.26 & 63.10 & \textbf{55.60} & 59.35 \\
        ZIG-2(or 5) & \textbf{24.33} & 37.24 & \textbf{28.51} & \textbf{17.03} & 27.00 & \textbf{59.50} & \textbf{62.30} & 50.40 & \textbf{38.29} $\uparrow$ & \textbf{2.88} & \textbf{2.02} & \textbf{2.45} & \textbf{63.80} & 55.40 & \textbf{59.60} \\
        \midrule
        FO-4(or 3)   & 23.22 & 37.31 & \textbf{28.77} & 16.15 & \textbf{27.20} & 58.37 & 59.00 & \textbf{51.80} & 37.71 & \textbf{2.82} & 1.88 & 2.35 & \textbf{65.80} & 52.40 & 59.10 \\
        ZIG-4(or 3)  & \textbf{23.46} & \textbf{37.67} & 28.75 & \textbf{16.20} & 27.00 & \textbf{60.05} & \textbf{60.80} & 50.67 & \textbf{38.08} $\uparrow$ & 2.64 & \textbf{2.20} & \textbf{2.42} & 64.63 & \textbf{54.20} & \textbf{59.42} \\

        \bottomrule
    \end{tabular}
\vspace{-0.2cm}
\end{table*}

\subsection{Guidance Analysis}
\label{sec:exp_guidance_analysis}

Refer to Appendix \ref{appendix:guidance_analysis_setup} for setup of this section.

\subsubsection{G1: Boundary Sharpening}
\label{sec:exp_g1}

\noindent\textbf{Results.} 
For the pre-training setting, as demonstrated in Table \ref{tab:merged_results_g1}, we derive two observations:
(1) \textit{High-scoring data at the end optimizes performance.}
Regardless of the initial phase, configurations ending with high-scoring samples (e.g., SEG(l10-h10)) consistently yield significant gains. 
Conversely, concluding with low-scoring data typically leads to performance degradation (e.g., SEG(h90)). 
%
%
(2) \textit{Initialization with high-score data alone yields negligible benefits.}
Configurations that exclusively prioritize high scores at the start (e.g., SEG(h10)) perform comparably to the random baseline.
This may be attributed to the constraints of a fixed data volume, where selecting high-scoring samples early on inevitably leaves lower-quality data for the final stages.

For the SFT setting, the superior performance is achieved when both the beginning and the end of the training utilize high-scoring data (SEG(h10-h10)).
This stems from the fact that starting with high-scoring data stabilizes the transition from pre-training, while ending with high-scoring data consistently optimizes final performance.

\begin{figure}[!htb]
  \centering
  \includegraphics[width=0.98\columnwidth]{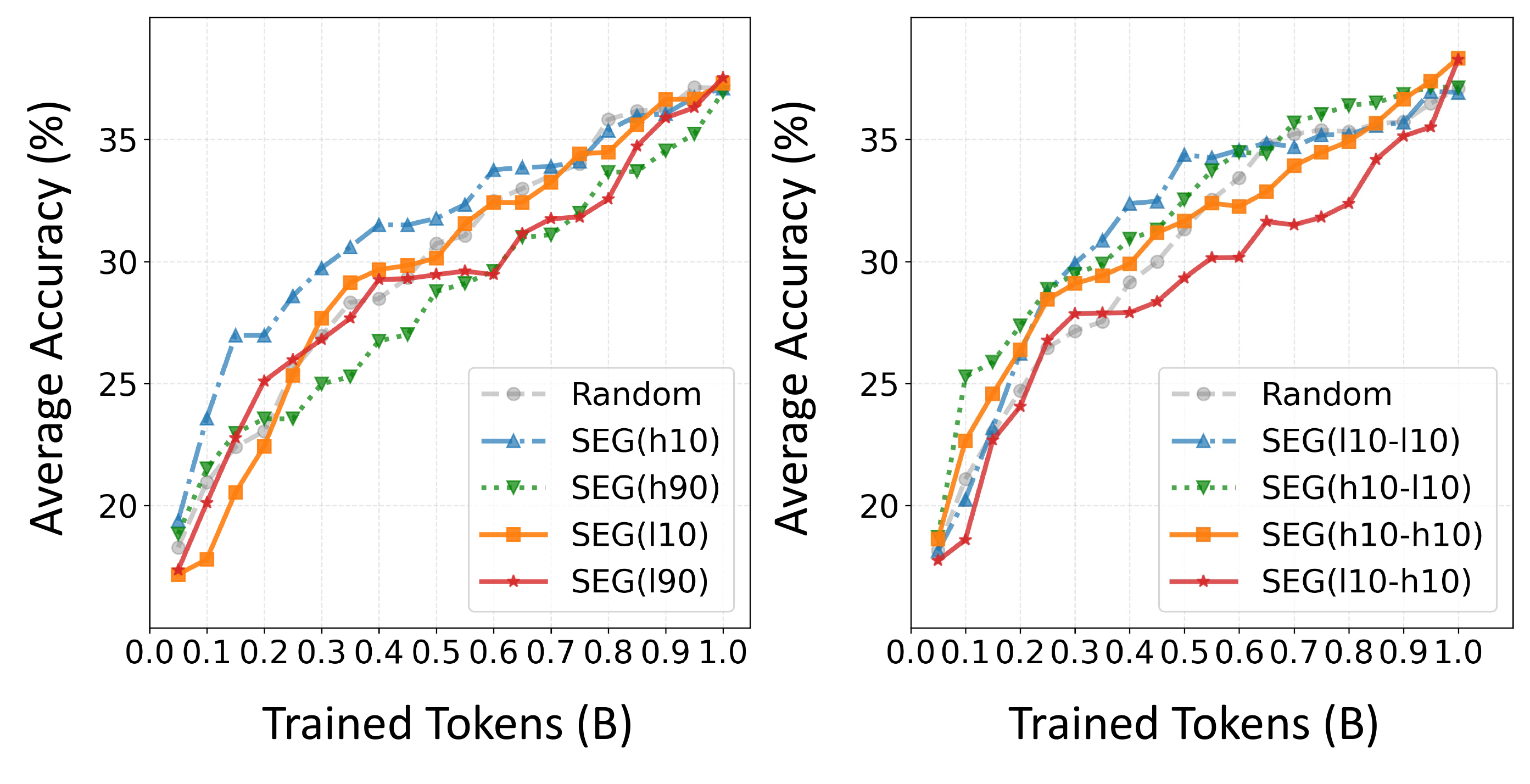}
    \vspace{-0.2cm}
    \caption{Comparison of accuracy trajectories for SEG. Results on Mistral-160M trained on 1B-tokens data.}
  \label{fig:acc_token_g1}
\vspace{-0.3cm}
\end{figure}

\noindent\textbf{Analysis.}
Figure \ref{fig:acc_token_g1} illustrates how data attributes at training boundaries shape model dynamic performance.
Variants ending with high-score data (e.g., SEG(l90), SEG(l10-h10)) exhibit a sharp final performance boost, surpassing the baseline.
%
In contrast, those ending with low-score data (e.g., SEG(h90), SEG(h10-l10)) suffer from stagnation.
%

\subsubsection{G2: Cyclic Scheduling}
\label{sec:exp_g2}

\noindent\textbf{Results.}
For the pre-training stage, as shown in Table \ref{tab:merged_results_guidance2}, both CL and FO methods achieve varying degrees of improvement over the random baseline.
The best variant (FO-2) consistently exceeds the CL baseline, proving that the periodic review of early foundational knowledge offers benefits to the learning process. 
For the SFT stage, 
FO yields substantial gains, especially with FO-3 in mathematics and FO-4/5 in coding. 
This confirms that review strategies remain effective in SFT stage.
However, the efficacy of FO across both stages is sensitive to the parameter $L$, as suboptimal selections can lead to performance degradation.



\begin{figure}[!tb]
  \centering
  \includegraphics[width=0.6\columnwidth]{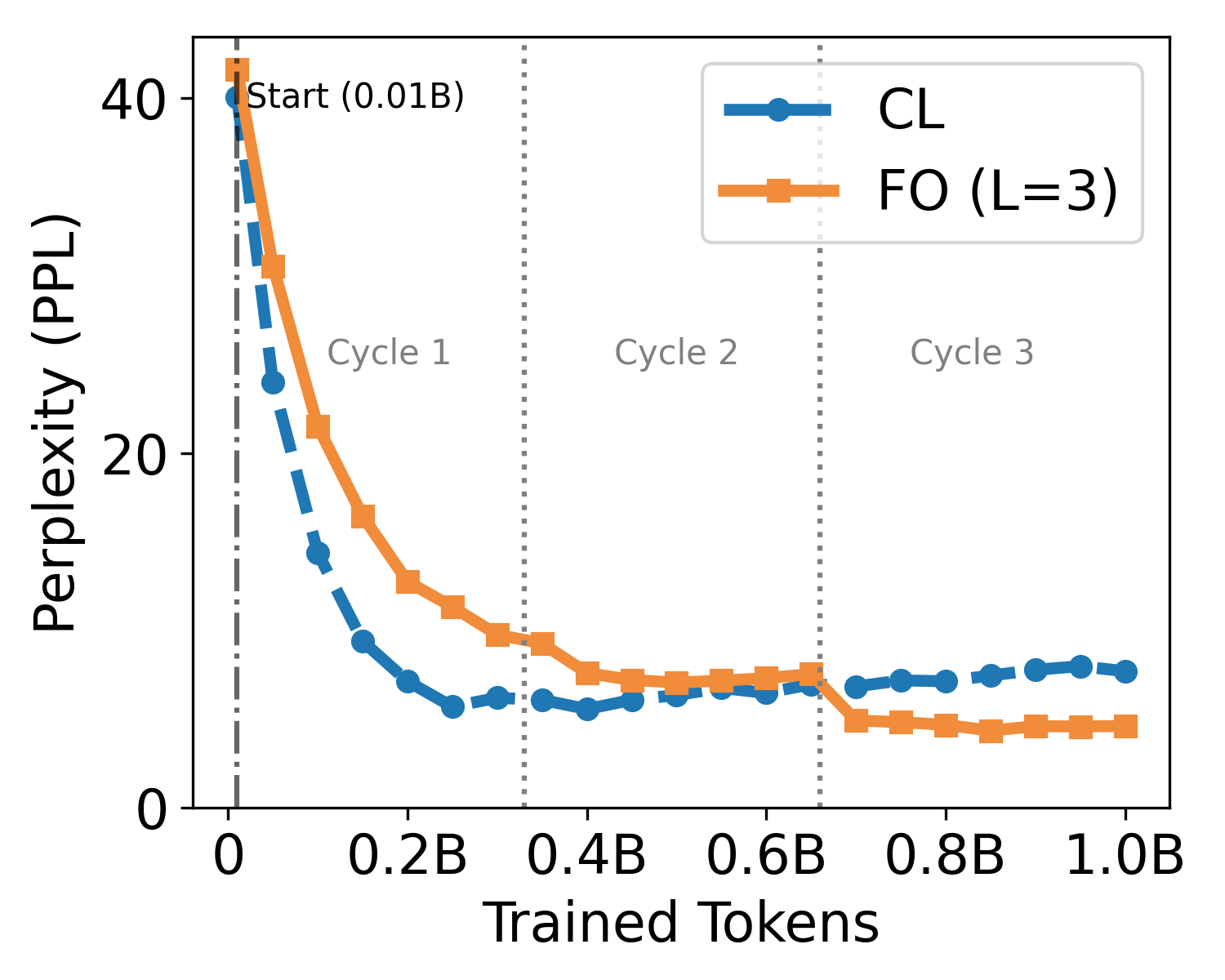}
\vspace{-0.2cm}
    \caption{The LMs’ perplexity (PPL) for $D_e$. Results on Mistral-160M trained on 1B-tokens data.}
  \label{fig:ppl}
\vspace{-0.6cm}
\end{figure}

\noindent\textbf{Analysis.}
Figure \ref{fig:ppl} presents the training-wise Perplexity (PPL) trends measured on the data with the 10\% lowest score, referred to as $D_{e}$, within $D_{\text{sort}}$.
The PPL of CL on $D_{e}$ rapidly drops (initial 30\% phase) but rebounds after entering the high-score data region (latter 50\%), which directly confirms the forgetting of early samples. 
For the FO-3, PPL drops normally in cycle 1. 
When simple data is reintroduced in cycle 2, PPL shows a secondary sharp drop. 
By the end of training (cycle 3), the model maintains a low PPL on $D_{e}$ and exhibits no rebound phenomenon seen in CL.

\subsubsection{G3: Curriculum Continuity}
\label{sec:exp_g3}

\noindent
\textbf{Results.}
In Table \ref{tab:merged_results_guidance3}, for the pre-training stage, the ZIG variants consistently outperform their FO counterparts. 
For the SFT stage, ZIG remains comparable to or outperforms FO in most cases. 
These results indicate that abrupt shifts in data attributes can negatively impact training, while enhancing attribute continuity effectively mitigates such issues.

\begin{figure}[!htb]
  \centering
  \includegraphics[width=0.57\columnwidth]{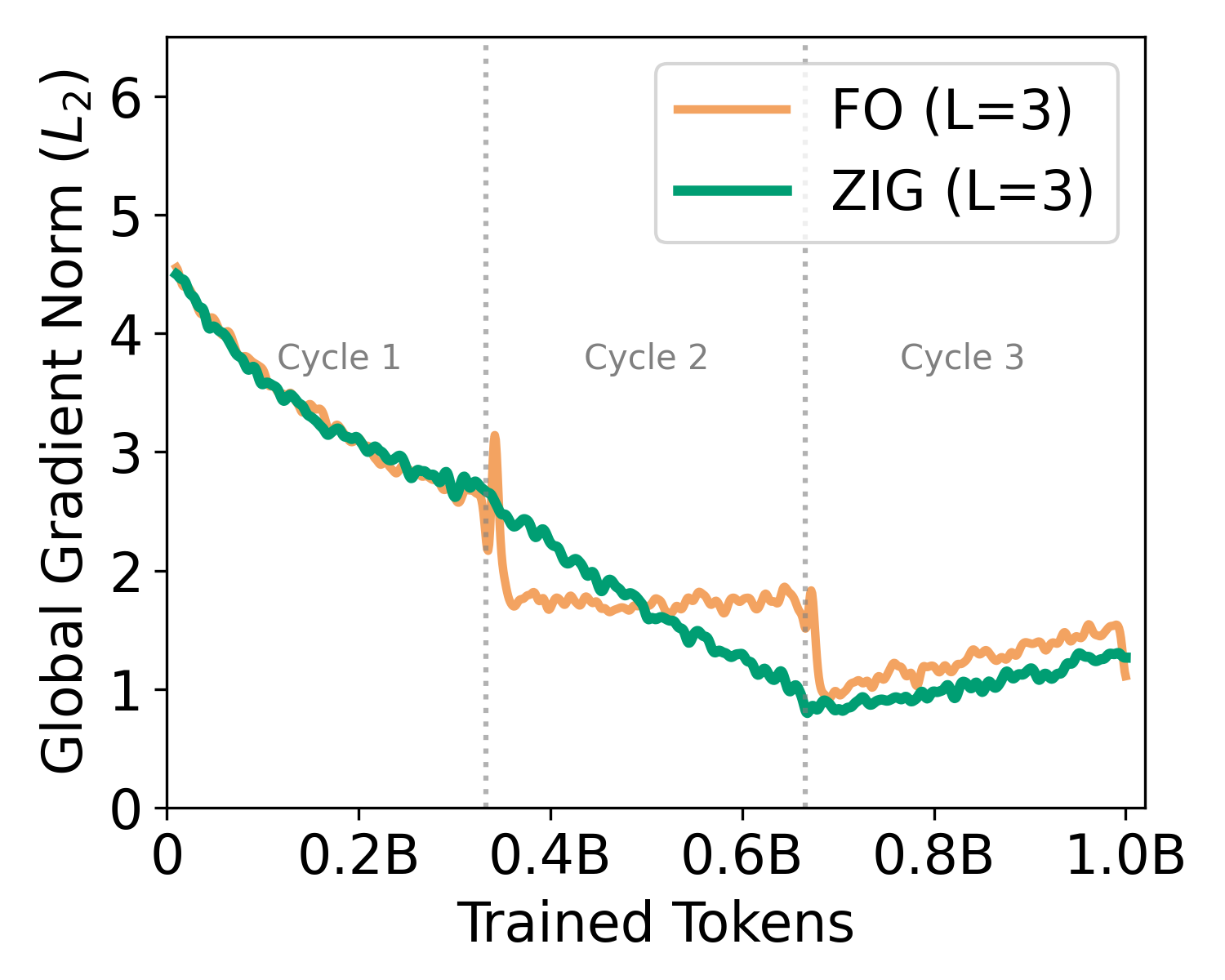}
\vspace{-0.2cm}
    \caption{Comparison of gradient norm trajectories. Results on Mistral-160M trained on 1B-tokens data.}
  \label{fig:gradient_norm}
\vspace{-0.4cm}
\end{figure}

\noindent
\textbf{Analysis.}
To empirically investigate the impact of abrupt data attribute shifts on optimization stability, we visualize the global gradient norm.
As shown in Figure \ref{fig:gradient_norm}, FO-3 exhibits a sharp spike in gradient norm at the boundary between cycles, indicating a shock to the optimizer due to the abrupt transition from samples with high to low score. 
In contrast, ZIG maintains a stable gradient magnitude, verifying its ability to ensure smoother optimization dynamics and stabilize the training process.

\subsubsection{G4: Local Diversity} 
\label{sec:exp_g4}

\noindent
\textbf{Results.} 
For both pre-training and SFT stages, as indicated in Table \ref{tab:results_guidance4_pretrain}, the application of the JIT strategy leads to consistent performance gains across almost all strictly sorted ordering methods. 
This observation suggests that the lack of local diversity acts as an inherent limitation in model training, which can be effectively mitigated by the JIT strategy to further improve model performance.

\begin{table}[htp]
    \centering
    \scriptsize
    \setlength{\tabcolsep}{2.5pt}
    \caption{Performance comparison of JIT variants and baselines to evaluate G4.}
\vspace{-0.2cm}

    \centering
    \label{tab:results_guidance4_pretrain}
    \centering
    \begin{tabular}{l|c|c|c|c}
    \toprule
          & FineWeb-Edu             & QuRatedPajama           & DeepMath          & OpenCodeInstruct        \\
            \midrule
    Random & \formattable{37.09}{0.08} & \formattable{35.60}{0.16} & \formattable{1.30}{0.02} & \formattable{55.47}{0.22} \\
    \midrule
    CL & 37.61 & 36.12 & 1.78 & 58.30 \\
    CL (JIT) & \formattable{\textbf{38.20}}{0.14} & \formattable{\textbf{36.46}}{0.11} & \formattable{\textbf{1.78}}{0.03} & \formattable{\textbf{59.50}}{0.35} \\
    \midrule
    FO & 38.12 & 36.62 & 2.42 & 60.90 \\
    FO (JIT) & \formattable{\textbf{38.25}}{0.09} & \formattable{\textbf{36.85}}{0.09} & \formattable{\textbf{2.74}}{0.12} & \formattable{\textbf{60.96}}{0.14} \\
    \midrule
    ZIG & 38.29 & 36.74 & 2.69 & 60.11 \\
    ZIG (JIT) & \formattable{\textbf{38.32}}{0.12} & \formattable{\textbf{36.88}}{0.10} & \formattable{\textbf{2.76}}{0.08} & \formattable{\textbf{61.34}}{0.18} \\
    \bottomrule
    \end{tabular}
\vspace{-0.2cm}
\end{table}

\begin{figure}[!htb]
  \centering
  \includegraphics[width=0.9\columnwidth]{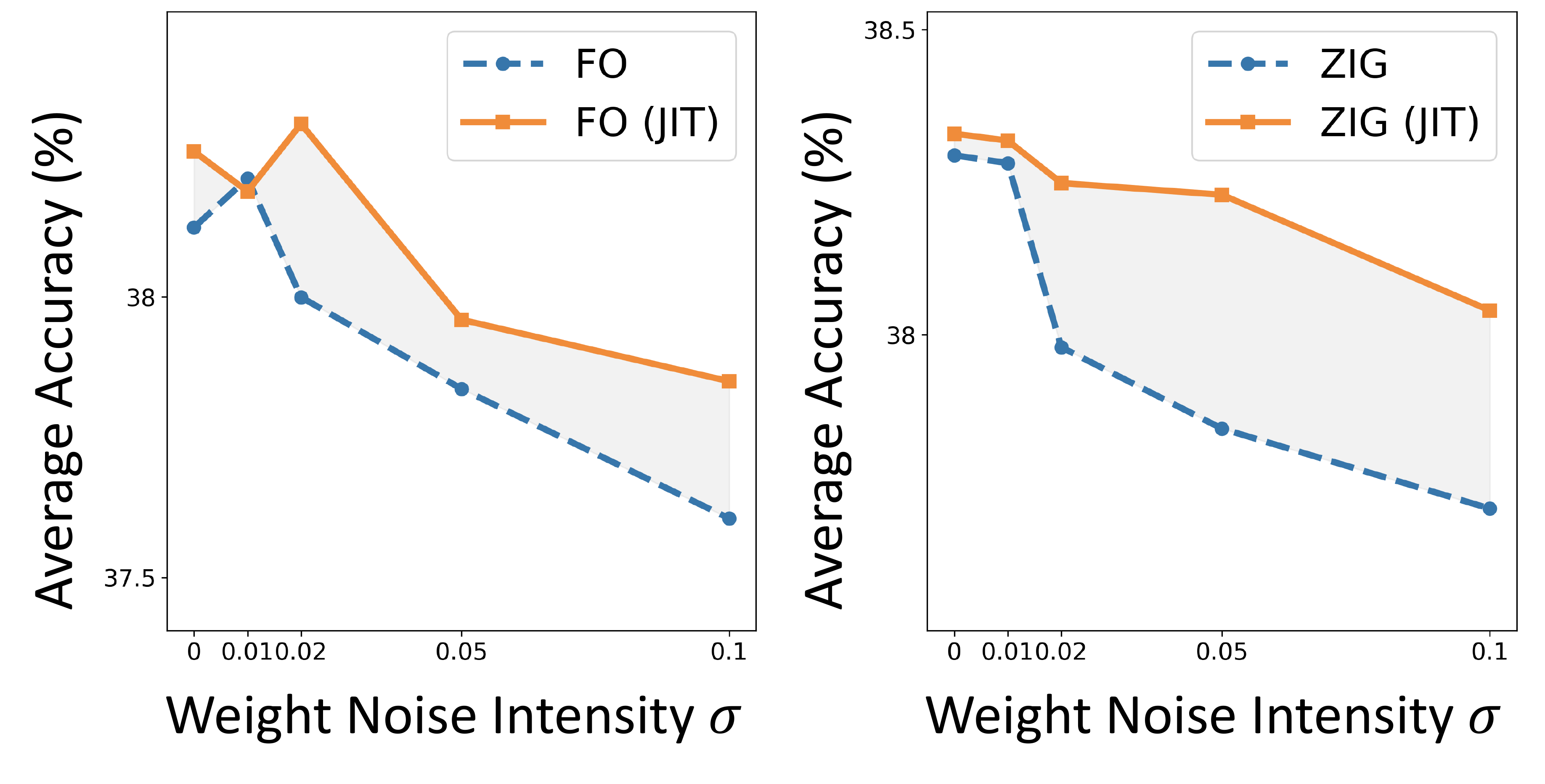}
\vspace{-0.2cm}
    \caption{Sensitivity analysis of model performance to weight perturbation.}
  \label{fig:flatness}
\vspace{-0.4cm}
\end{figure}

\noindent
\textbf{Analysis.}
Previous works have shown that the flatness of the loss landscape is strongly correlated with generalization and robustness \citep{keskar2016sharp,foret2020sharpness}. 
To evaluate how JIT enhances model robustness, we conduct weight perturbation analysis by injecting multiplicative Gaussian noise into parameters of various models (e.g., CL, FO, ZIG) and their JIT-enhanced version.

As illustrated in Figure \ref{fig:flatness}, the results show that  
(1) \textit{Superior performance.} Models with JIT consistently outperform baselines across most noise levels. 
(2) \textit{Flatter minima.} While the performance of baselines drops quickly as the noise scale $\alpha$ increases, JIT-enhanced models exhibit slower degradation. 
This lower sensitivity shows that JIT helps the model find broader and flatter minima, leading to better robustness and generalization potential.

\subsection{Main Results} 
\label{sec:exp_g1234}

\begin{table*}[htp]
    \centering
    \tiny
    \setlength{\tabcolsep}{2.5pt}

    \caption{Performance comparison of cross-guidance strategies (STR and SAW) against other strategies.
    24 and 25 refer to the AIME24 and AIME25, while HE refers to the HumanEval benchmark.
    }
    \label{tab:results_guidance1234}
    
        \begin{tabular}{l|cccc|cccccccc|c|ccc|ccc}
        \toprule
        & \multicolumn{4}{c|}{\textbf{Guidances}} & \multicolumn{9}{c|}{\textbf{Pre-training} (FineWeb-Edu)} & \multicolumn{6}{c}{\textbf{SFT} (DeepMath-103K | OpenCodeInstruct)} \\
        \midrule
               & G1 & G2 & G3 & G4 & ARC-c          & ARC-e          & HS             & LAMB          & OBQA          & PIQA           & SciQ          & Wino    & Avg.        & 24 & 25  & Avg. & HE & MBPP  & Avg.        \\
        \midrule
        Random & - & \checkmark & - & \checkmark & \formattable{21.47}{0.18} & \formattable{37.50}{0.23} & \formattable{27.57}{0.15} & \formattable{\textbf{15.97}}{0.12} & \formattable{25.60}{0.20} & \formattable{57.80}{0.51} & \formattable{59.50}{0.48} & \formattable{51.13}{0.04} & \formattable{37.09}{0.08} & \formattable{1.29}{0.02} & \formattable{1.30}{0.02} & \formattable{1.30}{0.01} & \formattable{57.93}{0.60} & \formattable{52.80}{0.06} & \formattable{55.37}{0.30} \\

        \midrule
        CL \citep{bengio2009cl}    & \checkmark & - & \checkmark & - & 23.11 & 38.63 & 28.10 & 13.85 & 28.40 & 57.42 & 60.10 & 51.50 & 37.61 & 1.94 & 1.61 & 1.78 & 61.59 & 55.00 & 58.30 \\
        DELT \citep{dai2025dataefficacy}     & \checkmark & \checkmark & - & - &  24.15 & 38.38          & 28.09          & 10.69          & \textbf{29.40} & 55.82          & 61.20 & 51.07   & 37.35       & 2.96 & 1.88 & 2.42 & 63.80 & 55.60 & 59.70          \\
        \midrule
        STR (Ours)    & \checkmark & \checkmark & - & \checkmark & 25.09 & 39.73 & \textbf{28.55} & 14.09 & 28.00 & \textbf{57.94} & 63.40 & \textbf{52.41} & 38.65 & \textbf{3.02} & 1.93 & 2.48 & \textbf{65.85} & 55.80 & \textbf{60.83} \\

        SAW (Ours)    & \checkmark & \checkmark & \checkmark & \checkmark & \textbf{25.37} & \textbf{39.86} & 28.44 & 14.65 & 29.00 & 57.72 & \textbf{63.50} & 51.64 & \textbf{38.78} & 2.91 & \textbf{2.14} & \textbf{2.53} & 64.76 & \textbf{56.20} & 60.48 \\

        \bottomrule
        \end{tabular}
\end{table*}

\textbf{Setup.}
To investigate the interplay among different guidances, we design the STR (Guidance 1\&2\&4) and SAW (Guidance 1\&2\&3\&4) strategies (see Sec. \ref{sec:cross-guidance}). 
%
Data organization methods of CL \citep{bengio2009cl} and DELT \citep{dai2025dataefficacy} are employed as baselines.
In the main text, we report the best performance across $L$ and JIT configurations.


\noindent
\textbf{Results.}
For both pre-training and SFT stages, as exhibited in Table \ref{tab:results_guidance1234}, the cross-guidance strategies of STR and SAW significantly outperform the CL and DELT baselines. 
Moreover, SAW and STR exhibit comparable performance. 
%
This parity likely stems from the fact that, unlike the full restart review in FO, STR narrows the scope of the review data, naturally mitigating the drastic attribute shifts. 
Therefore, the additional continuity benefits (Guidance 3) introduced by SAW become less apparent.
%

\subsection{Scaling-up Result}
\label{sec:exp_scaling_up}

\begin{table}[ht]
\centering
\scriptsize
\caption{
Scaling-up of different cross-guidance strategies during the pre-training stage on two 50B-token training datasets.
QR refers to the QuRatedPajama dataset, while FW refers the FineWeb-Edu dataset.
}
\vspace{-0.2cm}
\label{tab:results_scaleup_pretrain} 

\setlength{\tabcolsep}{3.5pt}
\begin{tabular}{l|cc|cc|cc|cc} 
\toprule
& \multicolumn{2}{c|}{160M} & \multicolumn{2}{c|}{470M} & \multicolumn{2}{c|}{1B} & \multicolumn{2}{c}{1.7B} \\
                & QR      & FW     & QR      & FW     & QR     & FW    & QR      & FW     \\
\midrule
Random          & 39.83              & 40.52           & 43.50              & 44.16           & 45.96             & 46.83          & 47.28              & 47.72           \\
\midrule
STR     & 41.27              & \textbf{43.13}           & 44.75              & \textbf{47.65}           & 46.33             & \textbf{48.45}          & \textbf{48.76}              & 49.85           \\
SAW     & \textbf{41.51}              & 43.10           & \textbf{45.07}              & 46.83           & \textbf{46.59}             & 48.06          & 48.42              & \textbf{50.11}           \\
\bottomrule
\end{tabular}
\vspace{-0.2cm}
\end{table}

\begin{figure*}[!tb]
  \centering
  \includegraphics[width=0.95\textwidth]
  {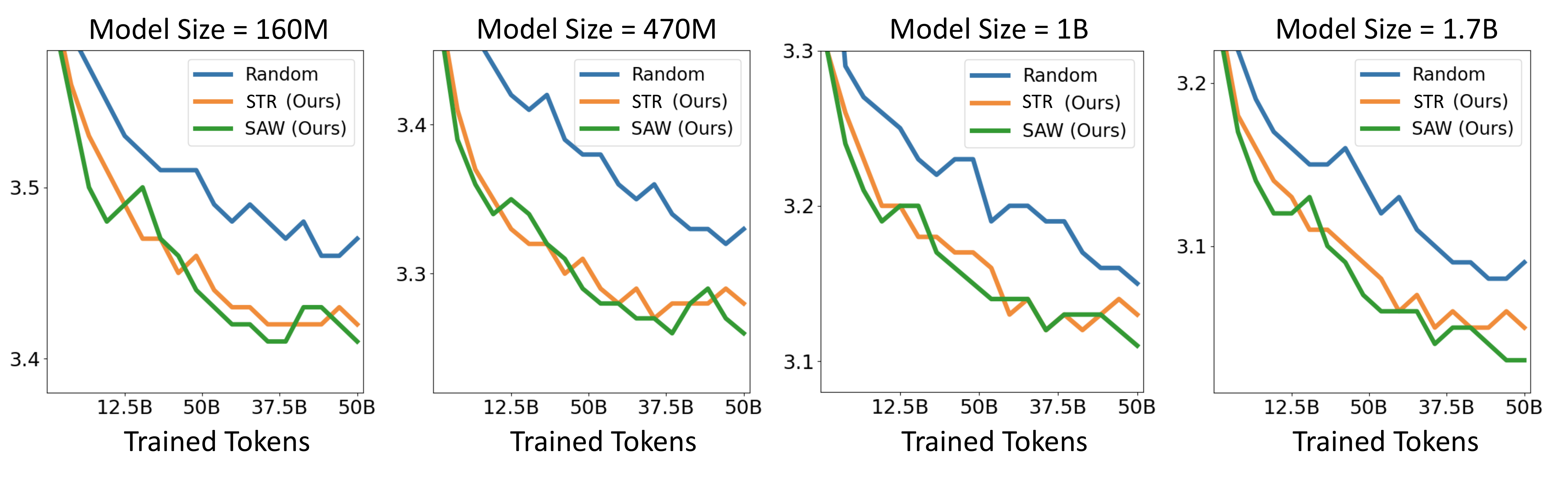}
    \caption{Test losses on the DCLM corpus \citep{dclm} across 160M to 1.7B model sizes. The labels STR and SAW refer to the optimal configurations: STR-2(JIT) and SAW-2(JIT).}
  \label{fig:dclm_loss}
\vspace{-0.3cm}
\end{figure*}

\begin{table}[t]
        \centering
        \scriptsize
        \makeatletter\def\@captype{table}\makeatother\caption{Test loss extrapolation via Chinchilla Scaling Law~\citep{chinchilla}. We estimate losses where model size $N$ and trained tokens $D$ align with GPT-3 175B, Llama 6.7B, Llama 2 70B, and Llama 3.1 405B. The improvements of data organization remain consistent for these LMs.}
        \vspace{-0.3cm}
        \begin{tabular}{l|rr|ccc}
            \toprule
                                    & $N$   & $D$   & Random      & STR & SAW          \\ \midrule
            GPT-3      & 175B  & 300B   & 2.876      & \textbf{2.859}  & \textbf{2.853} \\
            Llama     & 6.7B  & 1.0T   & 2.895      & \textbf{2.831}  & \textbf{2.824} \\
            Llama 2   & 70B   & 2.0T  & 2.788      & \textbf{2.759}  & \textbf{2.753} \\
            Llama 3.1 & 405B  & 15T   & 2.762      & \textbf{2.744}  & \textbf{2.738} \\
            \bottomrule
        \end{tabular}
        \label{tab:scaling_law_prediction}
\vspace{-0.5cm}
\end{table}

\textbf{Setup.} 
To assess the scalability and robustness of our ordering strategies, we evaluate our methods on a 50B-token data corpus across various Mistral model sizes, including 160M, 470M, 1B, and 1.7B parameters. 
We apply STR and SAW using the optimal configurations derived from the 160M/1B-token setup.
%
%

\noindent
\textbf{Results.}
As shown in Table \ref{tab:results_scaleup_pretrain}, both STR and SAW exhibit strong performance as model and data scales increase, consistently outperforming the random baseline across downstream tasks on average.
Beyond downstream tasks, we show that data organization also improves language modeling performance on high-quality corpora as the model scale increases.
Figure~\ref{fig:dclm_loss} illustrates the test loss on the DCLM subset, comparing conventionally pre-trained LMs (random) against those trained on data ordered by STR and SAW.
The results show that LMs trained on ordered data constantly achieve better performance across various model sizes. 

Table \ref{tab:scaling_law_prediction} details our extrapolation of test losses using the Chinchilla Scaling Law \citep{chinchilla}, showing that the gains from ordered data persist even when pre-training recent large LMs such as GPT-3 \citep{gpt3} and the Llama family \citep{llama,llama2,llama32024llama}.
More details of the extrapolation are available in the Appendix \ref{appendix:scaling_prediction}.
The DCLM corpus is assembled using a complex heuristic-based pipeline and is confirmed to be both diverse and exhaustive in its knowledge representation. 
Utilizing existing scores from data selection for data organization comes at a negligible cost and introduces a new perspective for enhancing model performance compared to complex curation pipelines.

\section{Conclusion}
This paper systematically explores the impact of data organization on Large Language Model (LLM) training, an area often overlooked despite its significance, especially given the limited number of training epochs for current LLMs.
By utilizing pre-computed sample-level scores from data efficiency methods, our approach introduces minimal computational overhead.
We identified and formalized four key guidances for optimizing training data organization.
According to them, two data ordering methods, STR and SAW, were proposed.
Extensive experiments consistently validated the effectiveness of these data ordering guidances and strategies enhance training stability and model performance across various model scales and data sizes.



\section*{Limitations}
While this work presents a systematic investigation into data organization for LLM training and introduces effective guidances, it is important to acknowledge certain limitations.
The methodology relies on reusing pre-computed sample-level scores, which, while offering the significant advantage of almost zero additional computational overhead, also means that the efficacy of the proposed organization strategies is contingent upon the quality and relevance of these pre-existing scores.


\section*{Ethics Statement}
We ensure transparency by utilizing publicly available data sources and model architectures.
We adhere to the licensing terms for both models and datasets, including proper attribution and sharing any modifications or derivative works under compatible terms.
While this work focuses on language modalities, further research is necessary for unbiased evaluation in other modalities.

\section*{Declaration of LLM Usage}
In the preparation of this work, the authors used LLM in order to improve the readability and language of the manuscript. After using this tool, the authors reviewed and edited the content as needed and take full responsibility for the content of the published article.

The data sources and model architectures applied in this work are publicly available to ensure transparency.
For both models and datasets used, we are committed to adhering to the terms of its license, which includes proper attribution ensuring that any modifications or derivative works are also shared under the compatible terms.
This work focuses on language modalities, and future work is needed for unbiased evaluation in other areas.

\bibliography{custom}

\appendix

\newpage

\section*{Appendix}
\label{sec:appendix}


\section{Extended Discussion on Guidances}

We offer additional insights and supplementary details regarding the proposed guidances to further clarify their mechanisms.

\begin{itemize}[left=0cm]
\item \textbf{Boundary Sharpening.}
Improper data characteristics at the start and end of training can hinder convergence \citep{2023boundary,kalra2024warmupboundary,paul2022lotteryboundary} or restrict peak performance \citep{hu2024minicpmboundary,llama32024llama}. 
We establish a boundary sharpening guidance, initiating training with low-score (simple, low-information) data to stabilize early optimization, while concluding with high-score (complex, high-quality) samples to align the model's capabilities with downstream reasoning tasks, thereby ensuring robust and performant models.
\item \textbf{Cyclic Scheduling.}
In a one-pass training regime \citep{tay2019simpleonepass,tudor2016hardonepass,wang2020curriculumonepass}, a strict easy-to-hard data curriculum often leads to the forgetting of early fundamental knowledge as the model transitions to complex samples \citep{wang2021clsurvey}. Moreover, the common practice of training LLMs for only one or a few epochs makes Baby Step methods \citep{bengio2009cl,spitkovsky2010babystep,cirik2016visualizingbabystep} impractical, given their requirement for replaying simple data whenever more challenging samples are introduced.
To address this, we propose cyclic scheduling guidance, which periodically feeds the model with data covering a broad score distribution.
%
\item \textbf{Curriculum Continuity.}
Abrupt fluctuations in data scores can shock the optimizer, causing loss divergence or training instability \citep{kong2021adaptivecontinuity,weinshall2018curriculumcontinuity}. 
We effectively maintain curriculum continuity, ensuring smooth transitions across the spectrum of sample attributes to allow the model to progressively adapt to evolving data patterns without disrupting the optimization momentum, which is crucial for maintaining stable learning trajectories.
\item \textbf{Local Diversity.} 
Strictly sorting data by score can reduce gradient diversity within mini-batches, thereby influencing the model to overfit specific patterns rather than learning generalizable features \citep{jiang2014selfdiversity,yin2018gradientdiversity}.
Consequently, the local diversity guidance suggests the introduction of diverse samples within each micro-batch while maintaining the overall curriculum progress, thereby maximizing the information entropy of gradient estimations and fostering the learning of generalizable features, ultimately improving model generalization.
\end{itemize}

\section{Related Work}
\subsection{Data Efficiency}
Data efficiency aims to optimize LLM training by identifying high-quality subsets from massive corpora, a guidance validated by state-of-the-art LLMs \citep{llama32024llama,openAI2024,qwen2025qwen3} which demonstrate that data quality often outweighs quantity.
Existing methods broadly fall into three categories: deduplication, distribution alignment, and quality-based scoring.
SemDeDup \citep{abbas2023semdedup} and D4 \citep{tirumala2023d4} focus on removing semantic redundancy to enhance diversity.
Beyond redundancy, DSIR \citep{xie2023data} selects subsets that mirror target distributions via importance weighting, while PDS \citep{gu2025data} evaluates sample utility based on gradient consistency.
More recently, fine-grained scoring systems have emerged to assess content quality; for instance, FineWeb-Edu \citep{penedo2023refinedweb} employs classifiers to identify educational content, and QuRating \citep{wettig2024qurating} evaluates text across multiple dimensions such as writing style and required expertise.
However, a critical inefficiency persists across these methods: while substantial computational resources are invested in deriving these scalar metrics, they are typically utilized solely for a one-off binary selection decision.
Once the subset is constructed, this rich scoring information is discarded, and the model trains on the retained data agnostic to the significant variance in their intrinsic quality or difficulty.

\subsection{Data Organization}
Data organization explores the strategic sequencing of training samples to optimize model convergence and performance.
It is distinguished from data efficiency, which focuses on subset selection.
The dominant paradigm is Curriculum Learning \citep{bengio2009cl}, which advocates for an easy-to-hard progression.
Various metrics have been proposed to quantify this difficulty, such as attention scores \citep{kim2024strategic} or soft edit distance \citep{chang2021does}.
Beyond monotonic sorting, advanced scheduling strategies have emerged: annealing approaches \citep{llama32024llama} prioritize high-quality data towards the conclusion of training to refine capabilities, while DELT introduces folding learning strategies to mitigate catastrophic forgetting by periodically exposing the model to the full difficulty spectrum \citep{dai2025dataefficacy}.
However, research on data organization remains fragmented, particularly in the context of LLMs, lacking guidances to guide the data ordering in model training.
To bridge this gap, our work is the first to systematically explore the guidances of data ordering, utilizing multi-dimensional scores derived from data efficiency (e.g., selection).
This approach allows us to construct sophisticated curricula that enhance model performance and training stability with almost no additional computational overhead.


\section{Additional Experimental Setup}
\label{appendix:setup_all}

\subsection{Setup for Guidances Analysis}
\label{appendix:guidance_analysis_setup}
\subsubsection{G1: Boundary Sharpening}
\textbf{Setup.}
To systematically investigate the impact of data distribution at the training boundaries, we use the SEG ordering strategy with diverse configurations (see Sec. \ref{sec:method_guidance1}).
We first isolate the effects of the initialization and ending phases by setting the segment parameter $L=2$.
Based on the percentile intervals of the sorted scores, we define four variants:
SEG(h10) ($\tilde{\mathcal{D}}_1 \in [0, 0.1]$, $\tilde{\mathcal{D}}_2 \in [0.1, 1]$);
SEG(l90) ($\tilde{\mathcal{D}}_1 \in [0.1, 1]$, $\tilde{\mathcal{D}}_2 \in [0, 0.1]$);
SEG(h90) ($\tilde{\mathcal{D}}_1 \in [0, 0.9]$, $\tilde{\mathcal{D}}_2 \in [0.9, 1]$);
and SEG(l10) ($\tilde{\mathcal{D}}_1 \in [0.9, 1]$, $\tilde{\mathcal{D}}_2 \in [0.1, 0.9]$).

Furthermore, to explore the joint influence of start-end distributions, we increase the granularity to $L=3$.
This setup includes the variants: 
SEG(h10-l10) ($\tilde{\mathcal{D}}_1 \in [0, 0.1]$, $\tilde{\mathcal{D}}_2 \in [0.1, 0.9]$, $\tilde{\mathcal{D}}_3 \in [0.9, 1]$) and its reverse SEG(l10-h10), 
as well as symmetric configurations SEG(l10-l10) ($\tilde{\mathcal{D}}_1, \tilde{\mathcal{D}}_3 \in [0.9, 1]$, $\tilde{\mathcal{D}}_2 \in [0, 0.9]$) 
and SEG(h10-h10) ($\tilde{\mathcal{D}}_1, \tilde{\mathcal{D}}_3 \in [0, 0.1]$, $\tilde{\mathcal{D}}_2 \in [0.1, 1]$).
Visualizations of the data distributions for each setting are provided in the Appendix.

\subsubsection{G2: Cyclic Scheduling}

\textbf{Setup.}
To thoroughly investigate the importance of knowledge review, we compare various configurations of CL and FO (see Sec. \ref{sec:method_guidance2}). 
For FO, we set $L \in \{2, 3, 4, 5, 20, 100\}$, based on optimal performance in preliminary folding-layer experiments.

\subsubsection{G3: Curriculum Continuity}

\textbf{Setup.}
To explore the impact of attribute continuity, we compare FO with its corresponding ZIG variants (see Sec. \ref{sec:method_guidance3}). 
%
Based on their performance of FO, we set our baselines using $L=2,3$ for pre-training and $L=3,5$ for SFT.
Identical settings are applied to ZIG for a consistent comparison.

\subsubsection{G4: Local Diversity}

\textbf{Setup.}
To evaluate the importance of local diversity, we integrate the JIT strategy (Sec. \ref{sec:method_guidance4}) into several data ordering variants that are strictly score-ordered, including CL, FO, ZIG. 
Table \ref{tab:results_guidance4_pretrain} presents the best-performing configuration among various JIT settings (as $w$ varies), where the window size $w$ is set to 5000, 50000, and 5000 for CL, FO, and ZIG, respectively.
%

\subsection{Data}
\label{appendix:setup data}
(1) \textbf{General data.}
To evaluate our method on general pre-training scenarios, we utilize two widely adopted corpora that provide comprehensive intrinsic sample-level scores for data selection.
\textbf{FineWeb-Edu} \citep{penedo2023refinedweb} is a large-scale dataset comprising 1.3T tokens sourced from CommonCrawl \citep{wenzek2020ccnet}. 
It employs a specialized classifier, distilled from annotations generated by Llama-3-70B, to assign an educational quality score (on a scale of 0 to 5) to each web page, prioritizing content with high knowledge density and logical reasoning.
\textbf{QuRatedPajama} \citep{wettig2024qurating} consists of 260B tokens derived from the CommonCrawl corpus. 
It distinguishes itself by offering fine-grained quality ratings across four specific dimensions: educational value, writing style, factual content, and required expertise.
For the main experiments (Sec. \ref{sec:exp_guidance_analysis} and \ref{sec:exp_g1234}), we randomly sample 1B tokens from the original corpus, while for the scaling experiments (Sec. \ref{sec:exp_scaling_up}), we sample 50B tokens.
Due to space limits, the main text focuses on FineWeb-Edu results while full QuRatedPajama data is deferred to the Appendix.

\textbf{Math data.}
We utilize \textbf{DeepMath-103K} \citep{he2025deepmath} to assess mathematical reasoning capabilities.
This dataset comprises approximately 103,000 instruction-tuning samples, constructed by augmenting standard benchmarks (e.g., GSM8K and MATH) with diverse reasoning paths using CoT strategies.
It provides a spectrum of difficulty based on the number of reasoning steps and logical complexity.

\textbf{Code Data.}
For programming tasks, we employ \textbf{OpenCodeInstruct} \citep{ahmad2025opencodeinstruct}, a large-scale code instruction dataset synthesized to enhance code generation and execution.
It evolves seed instructions into a broad range of algorithmic challenges, where samples are implicitly graded by their algorithmic complexity and execution-based correctness.

\subsection{Model}
\label{appendix:setup model}
For pre-training on general data, we adopt the same model architecture as Mistral~\citep{mistral}, pre-training variants with 160M, 470M, 1B, and 1.7B parameters.
Model configurations follow \citep{gu2025data}, with maximal learning rates following \citep{gpt3} detailed in Table~\ref{tab:model_config}.
For the main experiments (Sec. \ref{sec:exp_guidance_analysis} and \ref{sec:exp_g1234}), we apply the model with 160M parameters, while for the scaling experiments (Sec. \ref{sec:exp_scaling_up}), we apply models with 160M, 470M, 1B, and 1.7B parameters.

For the SFT stage, we initialize the model with Qwen3-1.7B-Base\footnote{\url{https://huggingface.co/Qwen/Qwen3-1.7B-Base}} and conduct full-parameter fine-tuning.

\begin{table}[h]
    \centering
    \scriptsize
    \begin{tabular}{l|cccccc}
    \toprule
    Model Size     & $d_{\text{model}}$ & $d_{\text{FFN}}$ & $n_{\text{layers}}$ & $n_{\text{head}}$ & $d_{\text{head}}$ & learning rate \\
     \midrule
    160M           &  768               &   3,072          & 12                  &  12               &  64               &  $6\times 10^{-4}$               \\ 
    470M           &  1,024             &   4,096          & 24                  &  16               &  64               &  $3\times 10^{-4}$               \\ 
    1B             &  1,536             &   6,144          & 24                  &  16               &  96               &  $2.5\times 10^{-4}$               \\ 
    1.7B           &  2,048             &   8,192          & 24                  &  16               &  128              &  $2\times 10^{-4}$               \\ 
     \bottomrule
    \end{tabular}
    \vspace{3mm}
    \caption{Model configurations and corresponding learning rates.}
    \label{tab:model_config}
\end{table}

\subsection{Training Configuration}
\label{appendix:setup training}
For pre-training on general data,
all models are trained with a batch size of 256 and a maximum input sequence length of 1,024 for one epoch. 
The AdamW optimizer \citep{adamw} is paired with a cosine learning rate scheduler. 
The scheduler includes a warm-up phase for the first 2,000 steps, after which the learning rate decays to 10\% of its peak value. 
The model architecture and corresponding learning rates are summarized in Table \ref{tab:model_config}.
The random baseline results are averaged over three independent runs with different seeds (8, 10, 42).

For SFT on mathematics and code datasets, we conduct our experiments using the Llama-Factory framework\footnote{\url{https://github.com/hiyouga/LlamaFactory}}.
The Qwen3-1.7B-Base model is trained using full parameter fine-tuning with a per-device batch size of 4 and 8 gradient accumulation steps for 3 epochs. 
The maximum input sequence length is set to 2,048, and sequence packing is enabled to improve computational efficiency. 
A cosine learning rate scheduler is employed with a peak learning rate of $2.0 \times 10^{-5}$. The scheduler includes a warm-up phase for the first 10\% of the training steps. All training is conducted in bfloat16 precision.

\textbf{Compute Resources.}
For the pre-training stage, we train the 160M model on 1B-token datasets using a single NVIDIA A100 GPU. 
For experiments with the 160M, 470M, 1B, and 1.7B models on 50B-tokens data, we utilize 8*A100 GPUs. 
For the SFT stage, all experiments are conducted on 4*A100 GPUs.
The random baseline results are averaged over three independent runs with different seeds (8, 10, 42).

\subsection{Evaluation}
\label{appendix:setup eval}
We evaluate the LMs' 0-shot accuracy on the downstream test datasets used in OLMo \citep{olmo}. 
We also report the LM's language modeling loss on a subset of DCLM \citep{dclm}.
For mathematics, we report results on AIME 2024 and 2025 \citep{aime} in the main text, with MATH-500 \citep{hendrycks2021math500} and Minerva Math \citep{lewkowycz2022minervamath} in the Appendix \ref{appendix:more_results}.
For code, we use HumanEval \citep{chen2021humaneval} and MBPP \citep{austin2021mbpp}.

%
For general scenarios, we assess the trained models on a range of standard natural language understanding and reasoning benchmarks, including Hellaswag (HS; \citealp{hella_swag}), Winogrande (Wino; \citealp{winogrande}), LAMBADA (LAMB; \citealp{lambada}), OpenbookQA (OBQA; \citealp{openbookqa}), ARC-easy/challenge (ARC-e/c;~\citealp{arc}), PIQA \citep{piqa}, SciQ \citep{sciq}, and BoolQ \citep{boolq}. 
We also evaluate the language modeling loss on a subset of DCLM \citep{dclm}, a high-quality curated dataset, to confirm that models trained on organized $D_\text{ord}$ preserve diversity and long-tail knowledge.
For the general benchmarks, the tasks are framed as multiple-choice questions, where the model selects the correct answer by minimizing the normalized loss across all candidate options (\texttt{acc\_norm}).

For domain-specific tasks, we assess the models on mathematical reasoning and code generation benchmarks. 
For mathematics, we report results on AIME 2024 and 2025 \citep{aime} in the main text, with MATH-500 \citep{hendrycks2021math500} and Minerva Math \citep{lewkowycz2022minervamath} deferred to the Appendix.
For code, we use HumanEval \citep{chen2021humaneval} and MBPP \citep{austin2021mbpp}.

For programming benchmarks, we evaluate the fine-tuned model's code generation capabilities on the HumanEval and MBPP benchmarks using their instruction-based variants. All inputs are formatted using the standard chat template to simulate user-assistant interactions. For HumanEval, we employ a zero-shot setting where the model is prompted to complete a function body based on its signature and docstring. The maximum generation length is set to 1,024 tokens. For MBPP, we adopt a 3-shot setting (\texttt{num\_fewshot=3}), where the exemplars are provided as multi-turn dialogue history to facilitate in-context learning. The maximum generation length for MBPP is constrained to 256 tokens. For both benchmarks, we utilize greedy decoding (\texttt{do\_sample=False}) and report the \texttt{Pass@1} metric to assess performance.

For the mathematical reasoning benchmarks, we adopt a process-generation, answer-evaluation paradigm. In this setup, the model is prompted with the raw problem text alongside four exemplar demonstrations (4-shot) that contain complete Chain-of-Thought reasoning, and is required to generate the full reasoning process before producing the final answer. We employ a stochastic decoding strategy with a temperature set to 0.7 and 
top-p set to 0.8 to encourage diverse reasoning paths. During evaluation, only the final answer is extracted and normalized for scoring.
To rigorously evaluate performance, we apply dataset-specific configurations: for the competition-level AIME 2024 and AIME 2025, we set the maximum generation length to 2048 tokens and generate k=64 independent samples per problem; for MATH-500 and Minerva Math, we set the maximum generation length to 1024 tokens and generate k=4 samples. 
The final performance is reported using the \texttt{avg@k} metric, which quantifies the model's accuracy across the sampled answers after standard normalization, reflecting the model's consistent performance over multiple reasoning attempts.

\section{Test Loss Extrapolation with the Scaling Law}
\label{appendix:scaling_prediction}

We extrapolate the test losses on the DCLM corpus~\citep{dclm} of the conventionally trained and PDS-trained LMs with the Scaling Law~\citep{chinchilla,scaling_law}. Following~\citet{chinchilla}, we consider the scaling law with the following form:
\begin{equation}
    L(N,D) = E + \frac{A}{N^\alpha} + \frac{B}{D^\beta},
    \label{eq:scaling_law}
\end{equation}
where $N$ is the model parameters, $D$ is the number of trained tokens, and $A$, $B$, $E$, $\alpha$, $\beta$ are constants. We obtain these constants by minimizing the Huber loss~\citep{huber_loss}:
\begin{equation}
\scriptsize
    \min_{a,b,e,\alpha,\beta} \sum_{(N_i, D_i, L_i)} \text{Huber}_{\delta}(\text{LSE}(a-\alpha \log N_i, b-\beta \log D_i, e) - \log L_i),
    \label{eq:scaling_law_loss}
\end{equation}
where $\text{LSE}(\cdot)$ is the log-sum-exp operation. The loss is summed over all $(N_i, D_i, L_i)$ tuples, which are obtained by the test losses of 160M, 470M, 1B, and 1.7B LM during training from 0B to 50B tokens. We record the losses every 2.5B tokens, resulting in a total $4 \times 50B/2.5B = 80$ tuples like $(N_i, D_i, L_i)$. After solving $a$, $b$, and $e$ from Eq. \ref{eq:scaling_law_loss} , we have $A=\exp(a)$, $B=\exp(b)$, and $E=\exp(e)$. 

\citet{chinchilla} optimizes Eq. \ref{eq:scaling_law_loss} using the LBFGS algorithm \citep{lbfgs}. However, we found this algorithm sensitive to the initialization of the parameters to be optimized. Therefore, we apply a two-stage optimization. Specifically, we first fit the following data scaling curves for $N=$160M, 470M, 1B, and 1.7B with non-linear least squares from \texttt{scipy.optimize.curve\_fit}\footnote{\url{https://docs.scipy.org/doc/scipy/reference/generated/scipy.optimize.curve_fit.html}}, which is much more robust to the initialization:
\begin{equation}
    L(D) = E'(N) + \frac{B_0(N)}{D^{\beta_0(N)}},
    \label{eq:data_scale}
\end{equation}
where $E'(N)$, $B_0(N)$ and $\beta_0(N)$ are the fitted parameters. Then, we fit the following model size scaling curve:
\begin{equation}
    E' = E_0 + \frac{A_0}{N^{\alpha_0}}.
    \label{eq:model_scale}
\end{equation}
We use the constants from Eq. \ref{eq:model_scale} and the average constants from Eq. \ref{eq:data_scale} to compute the initialization for the LBFGS algorithm:
\begin{equation}
\scriptsize
    \begin{aligned}
        a_0 &= \log A_0, \\
        b_0 &= \log \frac{B_0(\text{160M}) + B_0(\text{470M}) + B_0(\text{1B}) + B_0(\text{1.7B})}{4}, \\
        \alpha_0 &= \alpha_0, \\
        \beta_0 &= \frac{\beta_0(\text{160M}) + \beta_0(\text{470M}) + \beta_0(\text{1B}) + \beta_0(\text{1.7B})}{4}, \\
        e_0 &= \log E_0,
    \end{aligned}
\end{equation}
where $a_0, b_0, \alpha_0, \beta_0, e_0$ are the parameter initialization for the LFBGS algorithm to optimize Eq. \ref{eq:scaling_law_loss}. We set $\delta=1\times 10^{-3}$ and learning rate to $0.05$ when running LFBGS and obtain the constants in Table \ref{tab:scaling_constants}. We use these constants and Eq. \ref{eq:scaling_law} to compute the predicted loss in Table \ref{tab:scaling_law_prediction}.

\begin{table}[t]
    \centering
    \scriptsize
    \caption{Scaling law constants by fitting the test losses on the DCLM corpus. $E$ represents the irreducible loss inherent to the dataset.} 
    \label{tab:scaling_constants}
    \begin{tabular}{l|ccccc}
        \toprule
                    &  $A$                  &   $B$             &   $\alpha$    &   $\beta$     &   $E$      \\ \midrule
        Random      &  $4.82\times10^2$     &  $5.12\times10^3$ &   0.354        &   0.295        &   1.693    \\
        STR         &  $4.28\times10^2$     &  $4.55\times10^3$ &   0.359        &   0.299        &   1.693    \\
        SAW         &  $4.15\times10^2$     &  $4.38\times10^3$ &   0.361        &   0.302        &   1.693    \\
        \bottomrule
    \end{tabular}
\end{table}

\begin{table}[t]
    \centering
    \tiny
    \caption{Correlations ($R^2$) of fitting the scaling curves.}
    \label{tab:corr}
    \begin{tabular}{l|cccc|c}
    \toprule
                &  N=160M &   N=470M     &   N=1B    &   N=1.7B   & D=50$\times10^9$  \\   \midrule
        Random    &  0.992     &   0.995        &  0.998     &  0.997       & 0.999              \\
        STR       &  0.993     &   0.996        &  0.998     &  0.997       & 0.999              \\
        SAW       &  0.994     &   0.996        &  0.999     &  0.998       & 0.999              \\
    \bottomrule
    \end{tabular}
\end{table}

\paragraph{Goodness of Fit.} We evaluate the goodness of fit of the scaling curves with respect to the training token size $D$ and model size $N$ respectively, by computing the correlation coefficient $R^2=1-\frac{\sum_i(y_i-\hat{y}_i)^2}{\sum_i(y_i-\overline{y})^2}$, where $y_i$ is the ground truth value and $\hat{y}_i$ is the prediction. 

Regarding the training token size, to make the original problem a linear regression problem, we convert Eq. \ref{eq:scaling_law} into
\begin{equation}
\scriptsize
    \log \left(L(N,D)-E-\frac{A}{N^\alpha}\right) = \log B-\beta \log D.
\end{equation}
Then, we consider $\log \left(l_i-E-\frac{A}{N^\alpha}\right)$ as the ground truth value for regression, where $l_i$ is the observed loss, and $\log B-\beta \log D_i$ as the prediction. For each $N\in[160M, 470M,1B,1.7B]$ we compute an $R^2$ respectively. Similarly, regarding the model size, we convert Eq. \ref{eq:scaling_law} into
\begin{equation}
\scriptsize
    \log \left(L(N,D)-E-\frac{B}{D^\beta}\right) = \log A-\alpha \log N,
\end{equation}    
to compute the corresponding $R^2$ that measures its goodness of fit. For simplicity, we only consider the models at the end of training, where $D=50\times 10^9$. The results in Table~\ref{tab:corr} show that the correlations are sufficiently high, suggesting the scaling curve fits the impact from both the data and model sizes very well.




\section{Results on Additional Benchmarks}
\label{appendix:more_results}
In this section, we present additional experimental results to further validate our approach.For each guidance setting: 
(1) we conduct experiments on QuRatedPajama.
(2) we extend our evaluation to MATH-500 and Minerva Math. Note that the lower performance on AIME is primarily due to the inherent capacity constraints of the Qwen3-1.7B model.

Tables \ref{tab:appendix_merged_results_g1}, \ref{tab:appendix_merged_results_guidance2}, and \ref{tab:appendix_merged_results_guidance3} provide expanded results for G1, G2, and G3. 
Table \ref{tab:appendix_results_guidance1234} details additional findings for cross-guidance strategies. 
%
%


\begin{table*}[htp]
    \centering
    \scriptsize
    \setlength{\tabcolsep}{3.2pt}
    \caption{Performance comparison of SEG variants and baselines to evaluate G1.}
    \label{tab:appendix_merged_results_g1}
    
    \begin{tabular}{l | cccccccc |c | ccc }
        \toprule
        & \multicolumn{9}{c|}{\textbf{Pre-training} (QuRatedPajama)} & \multicolumn{3}{c}{\textbf{SFT} (DeepMath-103K)} \\
        \cmidrule(lr){2-10} \cmidrule(lr){11-13}
        & ARC-c & ARC-e & HS & LAMB & OBQA & PIQA & SciQ & Wino & Avg. & MATH500 & MinervaMath & Avg. \\
        \midrule
        Random & \formattable{21.08}{0.15} & \formattable{33.63}{0.20} & \formattable{27.50}{0.14} & \formattable{14.26}{0.10} & \formattable{25.80}{0.22} & \formattable{55.66}{0.45} & \formattable{55.50}{0.40} & \formattable{51.38}{0.42} & \formattable{35.60}{0.07} & \formattable{45.65}{0.62} & \formattable{9.28}{0.45} & \formattable{27.47}{0.12} \\
        \midrule
        SEG(h10)     & 21.76 & 33.08 & 27.04 & \textbf{15.99} & 25.00 & 55.55 & 53.80 & \textbf{52.09} & 35.54 & 46.20 & \textbf{12.50} & 29.35 \\
        SEG(h90)     & 21.59 & 30.72 & 27.52 & 12.73 & 23.20 & 55.06 & 53.30 & 49.57 & 34.21 & 47.80 & 11.21 & 29.51 \\
        SEG(l10)     & 20.56 & 33.42 & \textbf{27.81} & 15.89 & 26.00 & \textbf{55.82} & \textbf{58.80} & 50.36 & 36.08 & \textbf{48.55} & 11.12 & \textbf{29.84} \\
        SEG(l90)     & \textbf{23.29} & \textbf{36.07} & 27.29 & 13.47 & \textbf{28.00} & \textbf{55.82} & 55.80 & 51.93 & \textbf{36.46} & 48.15 & 11.12 & 29.64 \\
        \midrule
        SEG(h10-l10) & \textbf{22.44} & 30.51 & 27.29 & 13.25 & 24.40 & 54.13 & 54.50 & 49.72 & 34.53 & 44.25 & 9.10 & 26.67 \\
        SEG(l10-l10) & 21.67 & 30.43 & 27.33 & 14.21 & 23.60 & 55.44 & 53.50 & 51.46 & 34.70 & 45.65 & 10.66 & 28.16 \\
        SEG(l10-h10) & 22.01 & \textbf{36.57} & \textbf{27.71} & 14.09 & 25.20 & 55.88 & 59.50 & 51.30 & 36.53 & \textbf{49.45} & 11.58 & \textbf{30.52} \\
        SEG(h10-h10) & \textbf{22.44} & 35.31 & 27.54 & \textbf{14.28} & \textbf{26.00} & \textbf{57.07} & \textbf{60.20} & \textbf{51.78} & \textbf{36.83} & 46.35 & \textbf{13.24} & 29.79 \\
        \bottomrule
    \end{tabular}
\end{table*}

\begin{table*}[h]
    \centering
    \scriptsize
    \setlength{\tabcolsep}{3.5pt}
    \caption{Performance comparison of FO variants and baselines to evaluate G2.}
    \label{tab:appendix_merged_results_guidance2}
    
    \begin{tabular}{l | cccccccc|c | ccc }
        \toprule
        & \multicolumn{9}{c|}{\textbf{Pre-training} (QuRatedPajama)} & \multicolumn{3}{c}{\textbf{SFT} (DeepMath-103K)} \\
        \cmidrule(lr){2-10} \cmidrule(lr){11-13}
        & ARC-c & ARC-e & HS & LAMB & OBQA & PIQA & SciQ & Wino & Avg. & MATH500 & MinervaMath & Avg. \\
        \midrule
        Random & \formattable{21.08}{0.18} & \formattable{33.63}{0.23} & \formattable{27.50}{0.15} & \formattable{14.26}{0.12} & \formattable{25.80}{0.20} & \formattable{55.66}{0.51} & \formattable{55.50}{0.48} & \formattable{51.38}{0.04} & \formattable{35.60}{0.08} & \formattable{45.65}{0.62} & \formattable{9.28}{0.45} & \formattable{27.47}{0.12} \\
        CL     & 23.63 & \textbf{36.99} & 27.63 & 9.90 & 27.00 & 55.22 & 58.40 & 50.20 & 36.12 & 45.70 & 10.48 & 28.09 \\
        \midrule
        FO-2   & 23.29 & 36.57 & \textbf{27.96} & 11.14 & 27.00 & 56.04 & \textbf{59.60} & 51.38 & \textbf{36.62}  & 47.35 & 9.56 & 28.45 \\
        FO-3   & 23.12 & 35.77 & 27.71 & 11.78 & 25.40 & \textbf{56.31} & 57.20 & 50.04 & 35.92  & \textbf{50.25} & \textbf{13.13} & \textbf{31.69} \\
        FO-4   & 22.18 & 34.89 & 27.51 & 10.87 & \textbf{27.40} & 55.98 & 58.30 & \textbf{52.25} & 36.17  & 49.20 & 12.04 & 30.62 \\
        FO-5   & \textbf{23.72} & 34.18 & 27.72 & 11.95 & 25.80 & 54.13 & 56.10 & 52.17 & 35.72  & 44.55 & 10.48 & 27.51 \\
        FO-20  & 21.42 & 33.92 & 27.28 & 15.76 & 25.00 & 56.26 & 55.50 & 49.17 & 35.54  & 46.00 & 10.29 & 28.15 \\
        FO-100 & 21.25 & 34.05 & 27.78 & \textbf{16.20} & 26.20 & 56.04 & 55.70 & 50.51 & 35.97  & 45.05 & 11.86 & 28.45 \\
        \bottomrule
    \end{tabular}
\end{table*}

\begin{table*}[!h]
    \centering
    \scriptsize
    \setlength{\tabcolsep}{3.5pt}
    \caption{Performance comparison of ZIG variants and baselines to evaluate G3.}
    \label{tab:appendix_merged_results_guidance3}
    
\begin{tabular}{l | cccccccc | c | ccc }
        \toprule
        & \multicolumn{9}{c|}{\textbf{Pre-training} (QuRatedPajama)} & \multicolumn{3}{c}{\textbf{SFT} (DeepMath-103K)} \\
        \cmidrule(lr){2-10} \cmidrule(lr){11-13}
        & ARC-c & ARC-e & HS & LAMB & OBQA & PIQA & SciQ & Wino & Avg. & MATH500 & MinervaMath & Avg. \\
        \midrule
        Random & \formattable{21.08}{0.18} & \formattable{33.63}{0.23} & \formattable{27.50}{0.15} & \formattable{14.26}{0.12} & \formattable{25.80}{0.20} & \formattable{55.66}{0.51} & \formattable{55.50}{0.48} & \formattable{51.38}{0.04} & \formattable{35.60}{0.08} & \formattable{45.65}{0.62} & \formattable{9.28}{0.45} & \formattable{27.47}{0.12} \\
        \midrule
        FO-2 (or 4)   & 23.29 & 36.57 & 27.96 & 11.14 & 27.00 & 56.04 & \textbf{59.60} & 51.38 & 36.62 & 49.20 & 12.04 & 30.62 \\
        ZIG-2 (or 4)  & \textbf{23.35} & \textbf{36.73} & \textbf{29.21} & \textbf{11.52} & \textbf{27.40} & \textbf{56.60} & 58.10 & \textbf{52.07} & \textbf{36.87} & \textbf{49.80} & \textbf{12.83} & \textbf{31.32} \\
        \midrule
        FO-3   & 23.12 & 35.77 & 27.71 & \textbf{11.78} & 25.40 & 56.31 & \textbf{57.20} & 50.04 & 35.92 & 50.25 & \textbf{13.13} & \textbf{31.69} \\
        ZIG-3  & \textbf{23.44} & \textbf{36.01} & \textbf{27.96} & 11.64 &\textbf{26.80} & \textbf{56.77} & 56.70 & \textbf{52.02} & \textbf{36.42} & \textbf{50.65} & 12.57 & 31.61 \\
        \bottomrule
    \end{tabular}
\end{table*}

\begin{table*}[t]
    \centering
    \tiny
    \setlength{\tabcolsep}{2.5pt}

    \caption{Performance comparison of cross-guidance strategies (STR and SAW) against other strategies.
    }
    \label{tab:appendix_results_guidance1234}
    
        \begin{tabular}{l|cccc|cccccccc|c|ccc}
        \toprule
        & \multicolumn{4}{c|}{\textbf{Guidances}} & \multicolumn{9}{c|}{\textbf{Pre-training} (QuRatedPajama)} & \multicolumn{3}{c}{\textbf{SFT} (DeepMath-103K)} \\
        \midrule
               & G1 & G2 & G3 & G4 & ARC-c          & ARC-e          & HS             & LAMB          & OBQA          & PIQA           & SciQ          & Wino    & Avg.        & MATH500 & MinervaMath  & Avg.         \\
        \midrule
        Random & - & \checkmark & - & \checkmark & 21.08 & 33.63 & 27.50 & \textbf{14.26} & 25.80 & 55.66 & 55.50 & 51.38 & 35.60 & 45.65 &9.28 & 27.47 \\
        \midrule
        CL \citep{bengio2009cl}    & \checkmark & - & \checkmark & - & 23.63 & 36.99 & 27.63 & 9.90 & 27.00 & 55.22 & 58.40 & 50.20 & 36.12 & 45.70 & 10.48 & 28.09  \\ 
        DELT \citep{dai2025dataefficacy}     & \checkmark & \checkmark & - & -      & 23.29 & 36.57 & 27.96 & 11.14 & 27.00 & 56.04 & \textbf{59.60} & 51.38 & 36.62 & 49.20 & 12.04 & 30.62         \\
        \midrule
        STR (Ours)    & \checkmark & \checkmark & - & \checkmark &   \textbf{25.13}                & 37.17                & 28.13                & 12.25                & \textbf{27.30}                &\textbf{56.07}                & 57.30                & 52.00                & \textbf{36.92}       & 54.90 & 11.31 & 33.11         \\
        SAW (Ours)    & \checkmark & \checkmark & \checkmark & \checkmark & 24.49                & \textbf{37.34}                &\textbf{28.55}                & 12.28                & 27.20                & 56.04                & 56.40                & \textbf{52.14}                & 36.81  & \textbf{55.10} & \textbf{13.42} & \textbf{34.26}\\
        \bottomrule
        \end{tabular}
\end{table*}

\end{document}